\documentclass[10pt,twocolumn,letterpaper]{article}

\usepackage{iccv}              

\usepackage[accsupp]{axessibility}  
\usepackage{graphicx}
\usepackage{amsmath}
\usepackage{amssymb}
\usepackage{bbding}
\usepackage{booktabs}
\usepackage{extarrows}

\usepackage{indentfirst}
\usepackage{times}
\usepackage{epsfig}
\usepackage{color}
\usepackage{bm}
\usepackage{multirow}
\usepackage{makecell}
\usepackage{nicefrac}

\usepackage{mathrsfs}
\usepackage[table,xcdraw]{xcolor}
\usepackage[normalem]{ulem}
\usepackage{relsize}
\usepackage{tabularx}
\usepackage{algorithm}
\usepackage{algorithmic}
\usepackage[accsupp]{axessibility}
\usepackage{xcolor}
\definecolor{copynce}{RGB}{236,244,255}
\definecolor{baseline}{RGB}{232,232,232}

\newtheorem{coordinate_table}{Definition}
\useunder{\uline}{\ul}{}

\definecolor{iccvblue}{rgb}{0.21,0.49,0.74}
\usepackage[pagebackref,breaklinks,colorlinks,allcolors=iccvblue]{hyperref}

\def\paperID{7482} 
\def\confName{ICCV}
\def\confYear{2025}

\title{Tracing Copied Pixels and Regularizing Patch Affinity in Copy Detection}

\author{
   Yichen Lu
   \;\; Siwei Nie
   \;\; Minlong Lu
   \;\; Xudong Yang
   \;\; Xiaobo Zhang
   \;\; Peng Zhang\\
   Ant Group, China\\
   {\tt\small \{\href{mailto:luyichen.lyc@antgroup.com}{luyichen.lyc}, \href{mailto:niesiwei.nsw@antgroup.com}{niesiwei.nsw}, \href{mailto:luminlong.lml@antgroup.com}{luminlong.lml}, \href{mailto:jiegang.yxd@antgroup.com}{jiegang.yxd}, \href{mailto:ayou.zxb@antgroup.com}{ayou.zxb}, \href{mailto:minghua.zp@antgroup.com}{minghua.zp}\}@antgroup.com}
}

\begin{document}
\maketitle
\begin{abstract}


    Image Copy Detection (ICD) aims to identify manipulated content between image pairs through robust feature representation learning. While self-supervised learning (SSL) has advanced ICD systems, existing view-level contrastive methods struggle with sophisticated edits due to insufficient fine-grained correspondence learning. We address this limitation by exploiting the inherent geometric traceability in edited content through two key innovations. First, we propose PixTrace - a pixel coordinate tracking module that maintains explicit spatial mappings across editing transformations. Second, we introduce CopyNCE, a geometrically-guided contrastive loss that regularizes patch affinity using overlap ratios derived from PixTrace's verified mappings. Our method bridges pixel-level traceability with patch-level similarity learning, suppressing supervision noise in SSL training. Extensive experiments demonstrate not only state-of-the-art performance (88.7\% $\mu$AP / 83.9\% RP90 for matcher, 72.6\% $\mu$AP / 68.4\% RP90 for descriptor on DISC21 dataset) but also better interpretability over existing methods. Code is available\footnotemark[1].

\end{abstract}

\footnotetext[1]{https://github.com/eddielyc/CopyNCE}    
\section{Introduction}
\label{sec:intro}


Detecting manipulated multimedia content is crucial for various applications, including image/video retrieval systems and multimedia anti-piracy solutions~\cite{douze2009evaluation, Kim2003ContentbasedIC, Zhang2016ImageCD}. Significant research efforts have been devoted to addressing this challenge through image copy detection (ICD) and video copy detection (VCD) methods~\cite{giorgos2020learning, pizzi2022self, wang2022benchmark, He2022TransVCLAV, He2022ALC}, achieving impressive results. Self-supervised learning (SSL) has emerged as the predominant paradigm for training ICD models to learn representations effective for identifying exact duplicates, near-exact duplicates, and edited copies. However, current state-of-the-art approaches~\cite{pizzi2022self, yokoo2021contrastive, Wang2021BagOT} primarily employ view-level learning while neglecting correspondences at region- or patch-level granularity. We argue that this limitation hinders performance against sophisticated editing transformations.

Prior work in SSL has explored learning fine-grained correspondences through nearest-neighbor (NN) matching strategies~\cite{li2021esvit, wang2022exploring, Yun2022PatchlevelRL, Ziegler2022SelfSupervisedLO, Li2021DenseSC}. These approaches typically establish region-level correspondences by identifying either (1) feature-based NN matches or (2) location-based NN matches using patch centroid coordinates. However, the nearest-neighbor approach itself is not accurate enough and might leads to false matches (taking negative sample as positive and vice versa) or partial match (missing some positive samples). Applying such noisy supervision directly during training creates conflicting gradient signals that is detrimental to model convergence and final detection performance. This raises a question: How can we establish reliable pixel-wise correspondences between edited image pairs and effectively integrate this geometric awareness into SSL frameworks to enhance ICD robustness?

\begin{figure}[t]
    \begin{center}
        \includegraphics[width=1.0 \linewidth]{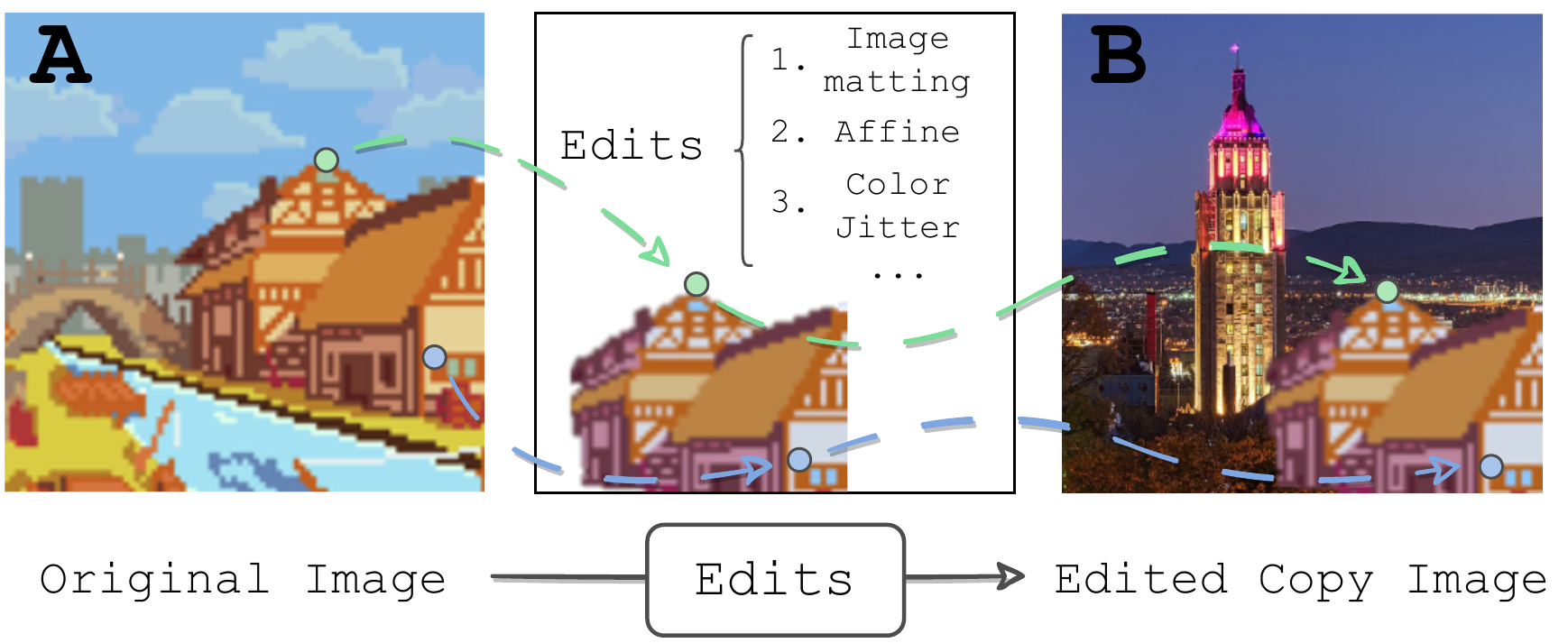}
    \end{center}
    \vspace{-0.4cm}
    \caption{\textbf{Traceability of pixels.} Image B copy edits upon image A. The edits includes image matting, affine transform and color jitter. Pixels of the copy region could be tracked back to the original image if specific edit functions are available.}
    \label{fig:traceability}
    \vspace{-0.4cm}
\end{figure}

A key property of manipulated content lies in its inherent geometric ``traceability'': \textit{Pixel correspondences between original and edited regions can be traced through sequential editing operations}. In principle, when considering standard SSL augmentation transforms~\cite{he2020moco, chen2020simclr, Caron2021EmergingPI} as edit operations, we should obtain fully traceable pixel relationships governed by deterministic transformation functions, as illustrated in Fig.~\ref{fig:traceability}. 

 Motivated by the inherent traceability of pixels after edits, we developed \textbf{PixTrace} which employs a coordinate table to track edited pixels. The table is sequentially updated by transform function associated with each editing operation, thereby consistently maintaining explicit coordinate mappings through editing sequences. By leveraging these precise pixel-level correspondences, our method eliminates ambiguous supervision from mismatched regions, thereby effectively suppressing noise from non-corresponding areas during contrastive learning.

Combined with \textbf{PixTrace}, we propose \textbf{CopyNCE}, a contrastive loss function that precisely aligns patch-level correspondences using geometrically-verified supervision. 
The loss regularizes patch affinity scores through a prior target distribution derived from the copy region overlap ratio between patch pairs. The overlap ratio can be estimated by \textbf{PixTrace}. As a result, supervision of CopyNCE will transfer pixel-level traceability to patch-level and encourage edited copy patches to identify each other.

\label{sec:contributions}
To conclude, our contributions could be summarized as:
\begin{itemize}
    \setlength{\itemsep}{0pt}
    \setlength{\parsep}{0pt}
    \setlength{\parskip}{0pt}
    \item[\textbf{i)}] We develop a comprehensive coordinate mapping pipeline \textbf{PixTrace} that maintains the traceability of edited pixels;
    \item[\textbf{ii)}] We propose \textbf{CopyNCE} that provides pixel-level guidance to regularize affinity between patches for better copy detection and localization;
    \item[\textbf{iii)}] CopyNCE series achieves 88.7\% $\mu$AP / 83.9\% RP90 for matcher and 72.6\% $\mu$AP / 68.4\% RP90 for descriptor on DISC21 dataset. Extensive experiments indicate that CopyNCE outperforms existing SOTAs not only in performance but also in interpretability and efficiency.
\end{itemize}

\section{Related Work}
\label{sec:related_work}

\subsection{Copy Detection}
    \label{sec:copy_detection}
    Copy detection aims to identify copy behavior in media content, \textit{e.g.} video~\cite{pizzi20242023, KordopatisZilos2023SelfSupervisedVS, He2022LearnFU, Han2021VideoSA} and image~\cite{Lee2024E2eViT, Wang2024PatternExpandableIC, Tan2024VisionTA, wang2024anypattern}. Self-supervised learning (SSL) has become the de facto paradigm for ICD/VCD. The typical way is building copy image pairs by various transforms. Although it has achieved impressive progress with contrastive learning~\cite{pizzi2022self}, metric learning~\cite{wang2022benchmark, Wang2021D2LVAD}, auxiliary token~\cite{Lu2024RTR}, fine-grained correspondences are not exploited. Model has to discovers copy regions with merely image-level annotations. A straightforward approach to mitigate this issue is adding an object detection model to crop portions of suspected copy regions or applying heuristic rules to extract multiple local regions for ensembling, as seen in leading DISC21~\cite{douze20212021} competition solutions~\cite{Wang2021D2LVAD, yokoo2021contrastive, Wang2021BagOT}. But these methods also face the risk of detection error propagation, and computational overhead from complicated pipeline.

    
\subsection{Local Visual Self-supervised Learning}
    \label{sec:visual_ssl}
    After success of self-supervised learning (SSL) in image level~\cite{he2020moco, chen2020simclr, Caron2021EmergingPI}, researches tend to explore SSL framework at finer granularity. These methods can be broadly categorized into four types. \textbf{\textit{(i).} Direct contrastive learning (CL) between features:} CL is conducted between features of different patches~\cite{Wang2020DenseCL} or pixels~\cite{VanGansbeke2021UnsupervisedSS}. 
    \textbf{\textit{(ii).} Building pseudo-positive local pairs with matching algorithms:} to dig better local features for CL, pseudo-positive local pairs could be identified with heuristic strategies, \textit{e.g.} nearest neighbors~\cite{li2021esvit, wang2022exploring, Yun2022PatchlevelRL}, clustering~\cite{Ziegler2022SelfSupervisedLO, Li2021DenseSC}, alignment algorithms~\cite{Xiao2021RegionSR, Huang2022LearningWT}, other models~\cite{ge2021revitalizing}, etc., which are neither accurate nor fine-grained enough.
    \textbf{\textit{(iii).} Leveraging human annotations:} to recall accurate positives, supervision from 
    segmentation~\cite{wen2022slotcon, Hu2021RegionawareCL, Henaff2021EfficientVP} or detection~\cite{Bai2022PointLevelRC} datasets can be based to discover regions with similar semantic content.
    \textbf{\textit{(iv).} Utilizing prior information in data pre-processing:} to ensure accuracy of positives without human annotations, augmentation, \textit{e.g.} overlay regions~\cite{Roh2021SpatiallyCR}, patch location ~\cite{bardes2022vicregl} and background replacement~\cite{Yang2021InstanceLF, Wang2022CP2} could provide external prior information. But its drawback of this type lies in overly simple augmentations and relatively coarse granularity. The primary issue when applied these methods in copy detection is that false matches or partial matches introduced by inaccurate correspondences, leading to suboptimal performance.

\subsection{Local Feature Matching}
    \label{sec:local_feature_matching}
    Local feature matching (LMF) obtains correspondence of keypoints between images, which contain different perspectives of the same object. A large body of works, \textit{e.g.} LoFTR~\cite{Sun2021LoFTRDL}, ASpanFormer~\cite{chen2022aspanformer} and CAPS~\cite{Wang2020LearningFD}, have reported effective solutions through camera pose and depth maps. Despite the success, depth maps can be incomplete and expansive to collect, and estimation error is also inevitable. To tackle this downside, SSL is introduced. It employs random homographic mapping on unlabeled images to yield the coordinates of keypoints before and after mapping. LMF model could be trained though regressing homographic matrix~\cite{DeTone2016DeepIH}, feature matching~\cite{DeTone2017SuperPointSI, Revaud2019R2D2RA}, etc. Inspired by earlier work, Glue series~\cite{Sarlin2019SuperGlueLF, Lindenberger2023LightGlueLF, jiang2024Omniglue} proposed to guide feature learning with pixel-level mappings and achieved SOTAs. Nevertheless, ICD necessitates complex editing that cannot be formulated by homographic matrix and it remains challenging to transfer keypoint-oriented loss directly to ICD.
    
\section{Approach}
\label{sec:approach}

\subsection{Background}
\label{sec:background}
    \begin{figure*}[h]
        \begin{center}
            \includegraphics[width=1.0 \linewidth]{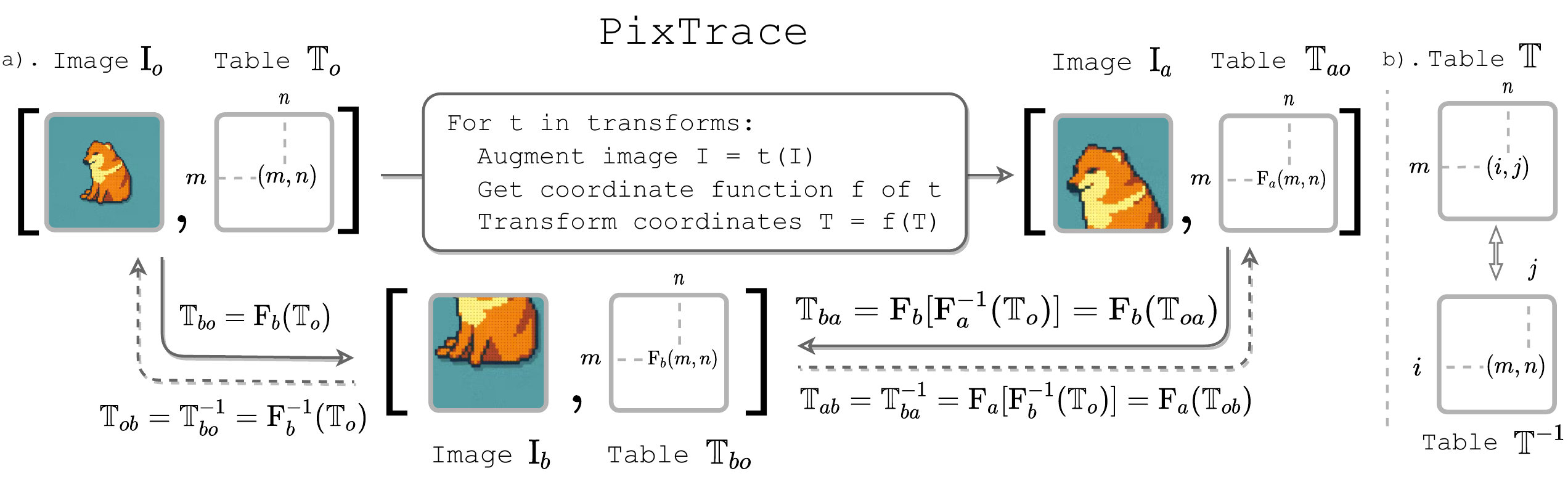}
        \end{center}
        \vspace{-0.4cm}
        \caption{\textbf{a). Overview of PixTrace.} Images $\text{I}_a$ and $\text{I}_b$ are both derived from $\text{I}_o$ through copy edits. Coordinate tables $\mathbb{T}_{ao}$ and $\mathbb{T}_{bo}$ describe coordinate correspondence of each pixel in $\text{I}_a$ and $\text{I}_b$ relative to $\text{I}_o$. Furthermore, with $\text{I}_o$ as a bridge, shared pixels in $\text{I}_a$ and $\text{I}_b$ can also be tracked against one another. Notice, edits in this figure are for demonstration purposes, actual edits are significantly more complex.
        \textbf{b). Illustration of reverse operation of table $\mathbb{T}$. } After reversion, coordinate $(i, j)$ at position $(m, n)$ will be used to place coordinate $(m, n)$.}
        \label{fig:pixtrace}
        \vspace{-0.2cm}
    \end{figure*}

    \noindent
    \textbf{Problem Definition.} In the context of image copy detection (ICD), we assume that image query $q$ is an edited copy of image reference $r$ or  $q$ and $r$ are edited copies of the same original image. Our goal is to obtain a model $f_\theta$ that can identify such pairs of images.

    \noindent
    \textbf{Descriptor \textit{v.s.} Matcher.} Generally, copy behaviors could be discovered through two forms of model: \textbf{descriptor} and \textbf{matcher}. Descriptor extracts features of images, represented as $z = f_\theta(x)$, and searches for edited copy images based on features cosine similarity. Matcher
    takes image pairs as input and directly performs binary classification. For simple practice, both descriptor and matcher are based on the Vision Transformer (ViT)~\cite{dosovitskiy2021an}. For descriptor, ViT serves as the encoder directly. For matcher, attention blocks function as both encoder and fusion modules, as illustrated in Fig.~\ref{fig:main}~(c). Detailed model structure is reported in Supp.

\subsection{PixTrace Pipeline}
    \label{sec:pixtrace}

    In SSL, most local modeling methods are heuristic, among which the approaches most closely related to PixTrace find patch correspondences based on features and locations. Feature-based methods (FeatNN)~\cite{li2021esvit, wang2022exploring, Yun2022PatchlevelRL} retrieve the nearest neighbor (NN) features as positives, while location-based methods (LocNN)~\cite{bardes2022vicregl} identify the corresponding locations of patch centroids of two views on their shared original image, selecting positives through location-based nearest patches. FeatNN is often disrupted by semantically similar regions and LocNN cannot ascertain whether patches overlap, which induce false match. Neither approach can determine the exact number of positive patches, which leads to partial match. Moreover varying importance of each positive patch is completely ignored by FeatNN and LocNN. These issues are of critical importance for copy detection. Typical bad cases are demonstrated in Fig.~\ref{fig:main}~(a).

    \begin{coordinate_table}
        \label{def:coordinate_table}
        Coordinate table $\mathbb{T}$ is a data structure based on dict that contains a collection of coordinate pairs, serving as keys and values. $\mathbb{T}$ supports following operations:
        \begin{itemize}
            \item $ \mathbb{T}[c] $ - Query the value with coordinate $c$ as the key.
            \item $\text{F}(\mathbb{T})$ - Update all values in $\mathbb{T}$ with function $\text{F}$.
            \item $\mathbb{T}^{-1}$ - Reverse key and value, \textit{i.e.,} mapping value to key.
        \end{itemize}
    \end{coordinate_table}

    To address these flaws, we construct a pixel tracking pipeline PixTrace to model pixel-wise coordinate mapping. In PixTrace, coordinate table $\mathbb{T}$ is defined as Def.~\ref{def:coordinate_table} and we expect $\mathbb{T}_{ao}$ to map coordinates of pixels in copy image $\text{I}_a$ to original image $\text{I}_o$. With table $\mathbb{T}$, all edited pixels are trackable, which aligns with the motivation behind CopyNCE.

    As illustrated in Fig.~\ref{fig:pixtrace}~(a), we perform image edits to imitate copy behavior. To construct an copy pair and its table $\mathbb{T}$, we take image $\text{I}_o$ and initialize $ \mathbb{T}=\mathbb{T}_o $, where each coordinate maps to itself, \textit{i.e.,} $ \mathbb{T}[m, n]=[m, n]$. Then, a sequence of edits (\textit{e.g.} affine, perspective, image matting, \textit{etc.}) is cast on image $ \text{I}_o $ to yield image $\text{I}_a $ and for each edit, we provide a function $f$ to perform coordinate transform to maintain the table. For simplicity, we combine functions of all edits into function $\text{F}_a = f_N \circ \cdots \circ f_1 $ and produce table $\mathbb{T}_{ao} = \text{F}_a(\mathbb{T}_o) $ and its reversed table $\mathbb{T}_{oa} = \mathbb{T}_{ao}^{-1} = \text{F}_{a}^{-1}( \mathbb{T}_o ) $. Note that reverse operation illustrated in Fig.~\ref{fig:pixtrace}~(b), could be considered a discrete representation of $\text{F}^{-1}$ at the pixel level and the reversed mapping can be directly obtained by $T^{-1}$ without specific $F^{-1}$. With the help of reverse operation, if another image $ \text{I}_b $ edits upon $\text{I}_o$ with its table $\mathbb{T}_{bo}$, then $\text{I}_o$ could bridge the pixel tracking between $\text{I}_a$ and $\text{I}_b$, \textit{i.e.,} $\mathbb{T}_{ba} = \text{F}_b [ \text{F}_{a}^{-1} ( \mathbb{T}_o ) ]$ and $\mathbb{T}_{ab} = \mathbb{T}_{ba}^{-1}$.

    Given PixTrace above, we have capability to track pixels between copy and original images, even between images containing content from the same original image. Now we can move on to CopyNCE.

\subsection{CopyNCE Loss}
    \begin{figure*}[h]
        \begin{center}
            \includegraphics[width=1.0 \linewidth]{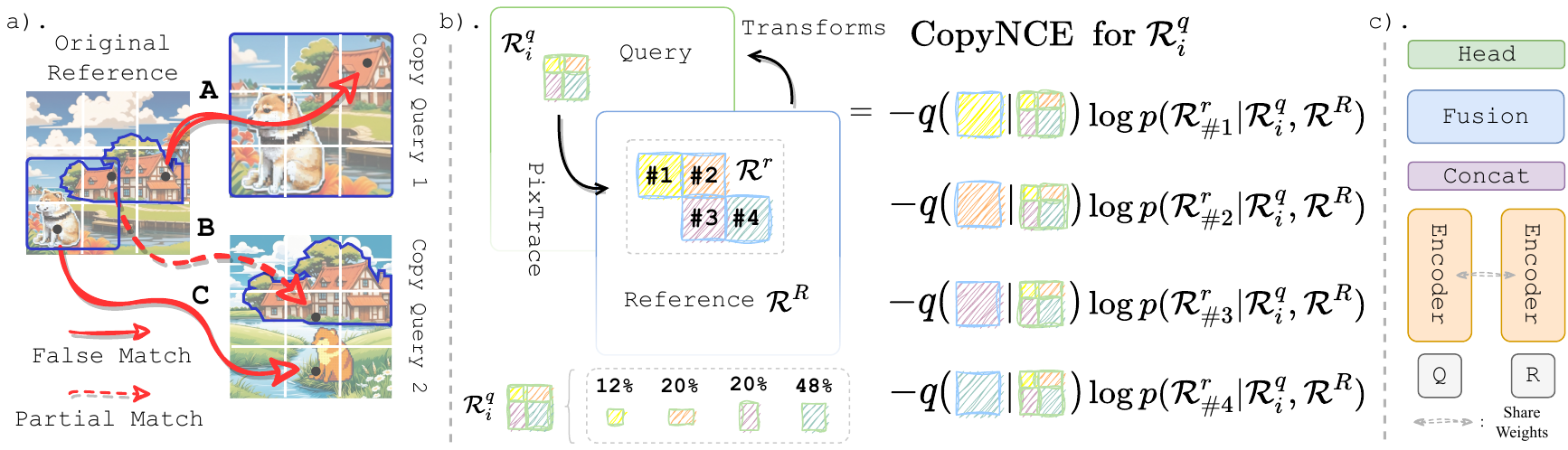}
        \end{center}
        \vspace{-0.4cm}
        \caption{\textbf{a). Typical noise of heuristic matching methods.} Match A: despite the nearest patch on query, LocNN still brings false match without overlapping since the original patch has no counterpart on Query 1; Match B: Neither LocNN nor FeatNN could retrieve all positive patches; Match C: FeatNN could be misled by semantically similar objects. \textbf{b). Overview of CopyNCE}. Patch $\mathcal{R}_i^q$ in query edits upon patches in $\mathcal{R}^r$ and their area proportions in $\mathcal{R}_i^q$ are \textcolor[RGB]{225, 225, 0}{12\%}, \textcolor[RGB]{255, 181, 112}{20\%}, \textcolor[RGB]{205, 162, 190}{20\%} and \textcolor[RGB]{154, 199, 191}{48\%}.  \textbf{c). Matcher Architecture.} Matcher consists of encoder and fusion modules. Fusion takes concatenated tokens from encoder as input. Encoders in matcher share weights. }
        \label{fig:main}
        \vspace{-0.2cm}
    \end{figure*}
    
    \label{sec:copynce}
    Since CopyNCE is inspired by InfoNCE, we first revisit the foundational InfoNCE proposed in CPC~\cite{Oord2018RepresentationLW}. InfoNCE aims to maximizes the mutual information $I(x; x_p)$ between a sample $x$ and its positive pair $x_p$. As demonstrated by Oord et al., this mutual information satisfies the lower bound $ I(x; x_p) \geq \log | X | + \mathbb{E}_X [ \log p(x_p | X, x) ] $, where $p(x_p | X, x)$ represents the probability of correctly selecting the positive sample $x_p$ from a noise set $X$ conditioned on $x$ and $| X |$ denotes the cardinality of $X$. To maximize $I(x; x_p)$, InfoNCE is formulated as follows:
    \begin{equation}
    \begin{aligned}
        \mathcal{L}_{\text{InfoNCE}}(x, x_p, X) & =  -\log p(x_p | X, x) \\
        & = - \log \frac{g_\theta(x, x_p)}{ \sum_{x_j \in X}{g_\theta(x, x_j)} }.
        \label{equ:infonce}
    \end{aligned}
    \end{equation}
    Generally, $g_{\theta}(x_i, x_j) = \exp( \cos( z_i, z_j ) / \tau )$, where $z_i$ and $z_j$ are features of $x_i$ and $x_j$. 

    In the context of ICD, suppose that a query image $q$ is an edited copy of a reference image $r$, with their corresponding manipulated regions denoted as $\mathcal{R}^q$ and $\mathcal{R}^r$. To guide the model in localizing these copy-forged regions, we propose to maximize mutual information $ I(\mathcal{R}^q; \mathcal{R}^r) $ as an auxiliary learning objective. Aligning with InfoNCE, we construct a contrastive noise set $\mathcal{R}^X = \{ \mathcal{R}^r \} \cup \{ \mathcal{R}_i^x \}_{i=1}^{N-1}$ where \( \mathcal{R}^x_i \) represents distractor regions. The CopyNCE loss is then derived as follows:
    \begin{equation}
        \begin{aligned}
            \mathcal{L}_{\text{CopyNCE}}^{\#1} & = -\log p(\mathcal{R}^r | \mathcal{R}^X, \mathcal{R}^q ) \\ 
            & = -\log \frac{f_{\theta}(\mathcal{R}^q, \mathcal{R}^r)}{\sum_{\mathcal{R}^x \in \mathcal{R}^X} f_{\theta}(\mathcal{R}^q, \mathcal{R}^x) },
        \end{aligned}        
    \end{equation}
    where $ f_{\theta} $ denotes the model with parameter $ \theta $ that embeds regions or patches of images.
    
    In mainstream visual architectures (\textit{e.g.} ViT and CNNs), a common approach for extracting features from arbitrary regions relies on ROI pooling~\cite{girshick2015fastrcnn}. While functional, ROI pooling inherently sacrifices fine-grained spatial details, leading to features that are excessively coarse for ICD tasks. To overcome this limitation, we apply CopyNCE at finer granularity by decomposing $\mathcal{R}^q$ and $\mathcal{R}^r$ into $\mathcal{R}^q = \cup_i \mathcal{R}_i^q$ and $\mathcal{R}^r = \cup_j \mathcal{R}_j^r$, where $\mathcal{R}_i$ or $\mathcal{R}_j$ is a minimal unit patch defined by the model architecture (\textit{e.g.} 16×16 pixels in ViT). If patches $ \mathcal{R}_i^q $ correspond to patch $ \mathcal{R}_{i+}^r $, the noise set at patch level can then be reformulated as $ \mathcal{R}^X = \{ \mathcal{R}_{i+}^{r} \} \cup \{ \mathcal{R}_j^x \}_{j=1}^{N-1} $, which turns CopyNCE into:
    \begin{equation}
        \begin{aligned}
            \mathcal{L}_{\text{CopyNCE}}^{\#2} & = \mathbb{E}_{\mathcal{R}_i^q} [ - \log p(\mathcal{R}_{i+}^{r} | \mathcal{R}^X, \mathcal{R}_i^q )] \\
            & = \mathbb{E}_{\mathcal{R}_i^q} [- \log \frac{g_{\theta}(\mathcal{R}_i^q, \mathcal{R}_{i+}^{r})}{\sum_{\mathcal{R}^x \in \mathcal{R}^X} g_{\theta}(\mathcal{R}_i^q, \mathcal{R}^x) } ].
            \label{equ:copynce_with_g}
        \end{aligned}
    \end{equation}
    In this formula, $ g_\theta(\mathcal{R}_i, \mathcal{R}_j) $ could be decomposed as $ \exp( \cos( z_i, z_j ) / \tau) $, where $ z_i = f_\theta(\mathcal{R}_i),  z_j = f_\theta(\mathcal{R}_j) $ denote corresponding tokens of specific patches encoded by model $ f_\theta $.
    
    It is noteworthy that each query patch $\mathcal{R}_i^q$ may exhibit multiple candidate correspondences $\mathcal{R}_{i+}^r$ in the reference image, with varying degrees of relevance quantified by their spatial overlap. In InfoNCE, we strive to make the probability of identifying the sole positive sample from the noise set as close to 1 as possible. With multiple positive patches, the probability of identifying various patches should be regularized by prior distribution $q(\mathcal{R}_j^r |  \mathcal{R}_i^q)$. We posit that the prior distribution can be defined by the area proportion of the shared region between patches. And in practice, we approximate area with pixel counts. Given a coordinate table $\mathbb{T}$ in Sec.~\ref{sec:pixtrace} to represent the correspondence between pixels before and after edits, we first model $\hat{q}(\mathcal{R}_j^r | \mathcal{R}_i^q)$ as:
    
    \begin{equation}
        \hat{q}(\mathcal{R}_j^r | \mathcal{R}_i^q) = \frac{| \{ \mathbb{T}[ c ] \in \mathcal{R}_j^r | c \in \mathcal{R}_i^q \} |}{ | \mathcal{R}_i^q | },
    \end{equation}
    where $ | \mathcal{R} | $ denotes number of pixels in region $\mathcal{R}$. To enhance the flexibility of the prior distribution, we introduce a confidence sharpening parameter $ \gamma $ that modulates the prior's certainty:
    \begin{equation}
        q(\mathcal{R}_j^r | \mathcal{R}_i^q) = \frac{\hat{q}(\mathcal{R}_j^r | \mathcal{R}_i^q)^{\gamma}}{\sum_{\mathcal{R}_{k}^{r} \in \mathcal{R}^r} \hat{q}( \mathcal{R}_{k}^{r} | \mathcal{R}_i^q )^{\gamma}}.
    \end{equation}
    The CopyNCE loss is further rewritten as:
    \begin{equation}
        \begin{aligned}
            & \;\;\;\; \mathcal{L}_{\text{CopyNCE}}^{\#3}(q, r, \mathbb{T}_{qr}) \\
            & = \mathbb{E}_{\mathcal{R}_i^q} \Big[ \sum_{\mathcal{R}_j^r \in \mathcal{R}^r}  q(\mathcal{R}_j^r, \mathcal{R}_i^q) \underbrace{\big[ - \log p( \mathcal{R}_j^r | \mathcal{R}^X, \mathcal{R}_i^q) \big]}_{\text{InfoNCE}}  \Big] \\
            & \xlongequal{ \mathcal{R}^X = \mathcal{R}^R } \mathbb{E}_{\mathcal{R}_i^q} \Big[ \sum_{\mathcal{R}_j^r \in \mathcal{R}^r} - q(\mathcal{R}_j^r, \mathcal{R}_i^q) \log p( \mathcal{R}_j^r | \mathcal{R}^R, \mathcal{R}_i^q)   \Big] \\
            & \equiv \mathbb{E}_{\mathcal{R}_i^q} \Big[ \text{KL}_{\mathcal{R}_j^r}[ q(\mathcal{R}_j^r, \mathcal{R}_i^q) \parallel p(\mathcal{R}_j^r | \mathcal{R}^R, \mathcal{R}_i^q)] \Big],
        \end{aligned}
        \label{equ:copynce_with_kl}
    \end{equation}
    where $\text{KL}[\cdot \parallel \cdot]$ denotes KL divergence. Guided by CPC~\cite{Oord2018RepresentationLW} requiring semantically hard negatives in InfoNCE, we construct $\mathcal{R}^X$ by collecting all patches from the reference image $ \mathcal{R}^X = \mathcal{R}^{R} $. This design intentionally retains in-image hard negatives: patches spatially adjacent to the positive sample $\mathcal{R}_{i+}^r$ often exhibit high visual similarity due to local texture continuity (\textit{e.g.}, repeated patterns in natural images), thereby posing challenging distractors for contrastive discrimination. Since the region correspondences are bidirectional, we derive the final symmetric form of CopyNCE as:
    \begin{equation}
        \mathcal{L}_{\text{CopyNCE}} = \frac{1}{2} \big[ \mathcal{L}_{\text{CopyNCE}}^{\#3}(q, r, \mathbb{T}_{qr}) + \mathcal{L}_{\text{CopyNCE}}^{\#3}(r, q, \mathbb{T}_{rq}) \big].
        \label{equ:copynce}
    \end{equation}
    Fig.~\ref{fig:main}~(b) illustrates the core idea of CopyNCE.

\subsection{Overall Objective}
    To integrate CopyNCE into the training pipeline and equip both descriptor and matcher with fundamental copy detection capabilities, CopyNCE will be combined with the baseline losses in the form of an auxiliary loss to constitute the final loss functions as follow:
    \begin{equation}
        \mathcal{L}^{\text{mat}} = \mathcal{L}_{\text{baseline}}^{\text{mat}} + w_{\text{NCE}} \mathcal{L}_{\text{CopyNCE}}, \\
        \label{equ:matching}
    \end{equation}
    \vspace{-0.6cm}
    \begin{equation}
        \mathcal{L}^{\text{des}} = \mathcal{L}_{\text{baseline}}^{\text{des}} + w_{\text{NCE}} \mathcal{L}_{\text{CopyNCE}}.
        \label{equ:descripotor}
    \end{equation}

\section{Experiments}
\label{sec:experiments}

    \subsection{Datasets and evaluation protocols}
        \label{sec:dataset_and_evaluation}
        \noindent
        \textbf{DISC21}~\cite{douze20212021} is introduced as the dataset of Image Similarity Challenge at NeurIPS’21 and has gained its popularity in ICD. DISC21 contains \texttt{training set} of 1M unlabeled images and \texttt{reference set} of 1M images. As query, \texttt{dev set} and \texttt{test set} have 50k images respectively. Note that query images in \texttt{test set} pose greater challenges than those in \texttt{dev set}. In addition, \texttt{dev set} is divided evenly into two halves. According to official rules, data in \texttt{dev set part I} is allowed to be used in finetuning. Both \texttt{test set} and \texttt{dev set part II} could serve for performance measurement.

        \noindent
        \textbf{NDEC}~\cite{wang2022benchmark} is a more challenging dataset built on basis of DISC21. To puts more emphasis on hard negatives, NDEC extends \texttt{training set} and \texttt{dev set part II} of DISC21 with hard negatives from OpenImage.
        
        \noindent
        \textbf{Evaluation Protocols.} For fair comparison, we adhere to common practices~\cite{pizzi2022self, wang2022benchmark, Wang2021D2LVAD}. Performance on DISC21~\cite{douze20212021} is evaluated on both $\mu$AP and RP90. $\mu$AP measures the model's capacity of image similarity ranking by calculating average precision (AP) across all image pairs. RP90 reports the recall rate for precision at 90\%. 

        \noindent
        \textit{More detailed discussion about datasets and evaluation protocols can be found in Supp.}

    \subsection{Implementation Details}
        \label{sec:implementation_details}
        \noindent
        \textbf{Training.} By default, both models are based on ViT-S pretrained by DINO~\cite{Caron2021EmergingPI} on ImageNet~\cite{deng2009imagenet} and optimized by AdamW. Learning rate starts from 1e-3 and 6e-4 for matcher and descriptor and scheduled by cosine. For matcher, CopyNCE is placed at 12-th layer, $ \tau = 1/16 $, $ w_{\text{NCE}} = 3 $, $ \gamma = 1 $. As for descriptor, CopyNCE is casted on 12-th layer and $ \tau = 1/16 $, $ w_{\text{NCE}} = 5 $, $ \gamma = 3 $. Default input size is $224 \times 224$. Hard negative mining and complex augmentation are employed. For simplicity, Binary Cross Entropy (BCE) acts as the matcher baseline and we follow practices in SSL adopt InfoNCE~\cite{Oord2018RepresentationLW} and KoLeo~\cite{sablayrolles2018spreading} loss as descriptor baseline. If not mentioned, no fine-tuning is performed on \texttt{dev set part I}.
        
        \noindent
        \textbf{Evaluation.} The default image input size is $224 \times 224$. If no additional statements are provided, no post-processing (\textit{e.g.} score normalization) or ensemble techniques (\textit{e.g.} local ensembling, multi-scale testing) are utilized.

        \noindent
        \textit{Due to sophisticated settings, all settings about training, finetuning and evaluation are listed in Supp.}

    \subsection{Comparison with Other SOTAs}
        \begin{table*}[h]
            \small
            \begin{center}
            \setlength{\tabcolsep}{1.35mm}
            \begin{tabular}{cccc|cc}
    
            \hline
            
            \hline\hline
            
            \multirow{2}{*}{Matcher} & \multicolumn{3}{c|}{Settings} & \multicolumn{2}{c}{Metrics} \\
            \cline{2-6}
    
            & Arch & Res. & Local & \makebox[0.04\textwidth][c]{$\mu$AP} & \makebox[0.04\textwidth][c]{RP90} \\
            \hline
    
            \multirow{3}{*}{Separate\ddag~\cite{Jeon20212ndPS}} & ViT-S & \multirow{3}{*}{$224 \times 112$}  & \XSolidBrush & 75.4 & 68.7 \\
            \cline{2-2}\cline{4-6}
    
            & ViT-B &  & \XSolidBrush & 78.4	& 72.9 \\
            \cline{2-2}\cline{4-6}
    
            & ViT-L &  & \XSolidBrush & 84.7	& 80.3 \\
            \hline
    
            \cellcolor[HTML]{ECF4FF}{\textbf{CopyNCE}} & \cellcolor[HTML]{ECF4FF}{\textbf{ViT-S}} & \cellcolor[HTML]{ECF4FF}{$224 \times 224$} & \cellcolor[HTML]{ECF4FF}{\XSolidBrush} & \cellcolor[HTML]{ECF4FF}{\textbf{83.5}} & \cellcolor[HTML]{ECF4FF}{\textbf{75.4}} \\

            \hline

            \cellcolor[HTML]{ECF4FF}{\textbf{CopyNCE}} & \cellcolor[HTML]{ECF4FF}{\textbf{ViT-S}} & \cellcolor[HTML]{ECF4FF}{$336 \times 336$} & \cellcolor[HTML]{ECF4FF}{\XSolidBrush} & \cellcolor[HTML]{ECF4FF}{\textbf{85.8}} & \cellcolor[HTML]{ECF4FF}{\textbf{79.9}} \\
    
            \hline

            \hline
            
            \hline
    
            ImgFp~\cite{Sun20213rdPA} & EsViT-B & $224 \times 224$ & \Checkmark & 61.2 & - \\
            \hline
            
            \multirow{3}{*}{Separate\ddag~\cite{Jeon20212ndPS}} & ViT-S & \multirow{3}{*}{$224 \times 112$} & \Checkmark & 77.1 & 70.5 \\
            \cline{2-2}\cline{4-6}
    
            & ViT-B &  & \Checkmark & 80.7 & 75.6 \\
            \cline{2-2}\cline{4-6}
    
            & ViT-L &  & \Checkmark & 86.2 & 82.2 \\
            \hline
                
            D$^2$LV~\cite{Wang2021D2LVAD} & Multi & $256 \times 256$ & \Checkmark & 88.6 & 80.1 \\
            
            \hline
    
            \cellcolor[HTML]{ECF4FF}{\textbf{CopyNCE}} & \cellcolor[HTML]{ECF4FF}{\textbf{ViT-S}} & \cellcolor[HTML]{ECF4FF}{$224 \times 224$} & \cellcolor[HTML]{ECF4FF}{\Checkmark} & \cellcolor[HTML]{ECF4FF}{\textbf{87.4}} & \cellcolor[HTML]{ECF4FF}{\textbf{81.3}} \\
    
            \hline
            
            \cellcolor[HTML]{ECF4FF}{\textbf{CopyNCE}} & \cellcolor[HTML]{ECF4FF}{\textbf{ViT-S}} & \cellcolor[HTML]{ECF4FF}{$336 \times 336$} & \cellcolor[HTML]{ECF4FF}{\Checkmark} & \cellcolor[HTML]{ECF4FF}{\textbf{88.7}} & \cellcolor[HTML]{ECF4FF}{\textbf{83.9}} \\
            
            \hline\hline
    
            \hline
            
            \end{tabular}
            \quad
            \begin{tabular}{cccc|cc}
    
            \hline
            
            \hline\hline
    
            \multirow{2}{*}{Descriptor} & \multicolumn{3}{c|}{Settings} & \multicolumn{2}{c}{Metrics} \\
            \cline{2-6}
    
            & Arch & Res. & Pre/Post & \makebox[0.04\textwidth][c]{$\mu$AP} & \makebox[0.04\textwidth][c]{RP90} \\
            \hline
    
            DINO~\cite{Caron2021EmergingPI} & ViT-S & $224 \times 224$ & \XSolidBrush & 20.0 & 6.8 \\
            \hline
    
            S-square\dag~\cite{Papadakis2021ProducingAE} & EffNet-B5 & $160 \times 160$ & \XSolidBrush & 66.4 & - \\
            \hline
    
            Lyakaap\dag~\cite{yokoo2021contrastive} & EffNetV2-M & $512 \times 512$ & \XSolidBrush & 64.3 & 56.6 \\
            \hline
    
            SSCD~\cite{pizzi2022self} & R50 & Long$ \times 288$ & \XSolidBrush & 61.5 & 38.3 \\
            \hline
            
            \cellcolor[HTML]{ECF4FF}{\textbf{CopyNCE}} & \cellcolor[HTML]{ECF4FF}{\textbf{ViT-S}} & \cellcolor[HTML]{ECF4FF}{$224 \times 224$} & \cellcolor[HTML]{ECF4FF}{\XSolidBrush} & \cellcolor[HTML]{ECF4FF}{\textbf{70.5}} & \cellcolor[HTML]{ECF4FF}{\textbf{63.6}} \\
            \hline

            \hline

            \hline

            \multirow{2}{*}{BoT~\cite{Wang2021BagOT}} & \multirow{2}{*}{R50} & \multirow{2}{*}{$224 \times 224$} & Str & 70.5 & 61.6 \\
            \cline{4-6}
    
            &  &  & YL / Str & 71.5 & 62.9 \\
            \hline

            SSCD~\cite{pizzi2022self} & R50 & Long$ \times 288$ & SN & 72.5 & 63.1 \\
            \hline
    
            \cellcolor[HTML]{ECF4FF}{\textbf{CopyNCE}} & \cellcolor[HTML]{ECF4FF}{\textbf{ViT-S}} & \cellcolor[HTML]{ECF4FF}{$224 \times 224$} & \cellcolor[HTML]{ECF4FF}{\textbf{SN}} & \cellcolor[HTML]{ECF4FF}{\textbf{72.6}} & \cellcolor[HTML]{ECF4FF}{\textbf{68.4}} \\
            
            \hline\hline
    
            \hline
            
            \end{tabular}
            
            \end{center}
            
            \vspace{-0.4cm}
            
            \caption{\textbf{Comparison with other SOTA methods.} \textbf{Left} is for matcher and \textbf{Right} is for descriptor. \textbf{Local} denotes inference ensembling with multiple local crops. \textbf{Pre/Post} is the pre-/post-processing, in which \textbf{SN} is score normalization, \textbf{YL} is YOLO pre-processing and \textbf{Str} is feature stretching. \dag \; denotes the method leverages extra data for training. \ddag \; means that we reproduce the results with its open-source code. \textbf{Multi} in D$^2$LV stands for 11$\times$R50~\cite{He2016DeepRL}, 11$\times$R152~\cite{He2016DeepRL} and 11$\times$R50IBN~\cite{pan2018ibn}.}  
            
            \label{tab:comparison_with_sotas}
            \vspace{-0.2cm}
        \end{table*}

        In ICD, both descriptor and matcher are crucial. Therefore, we conduct extensive experiments on both models. Note that, since most recent methods are competition solutions, settings vary significantly. Thus, we could only provide the most equitable available results.

        \noindent
        \textbf{Matcher.} As shown in the left part of Tab.~\ref{tab:comparison_with_sotas}, CopyNCE achieves 83.5\% $\mu$AP / 75.4\% RP90, without other bells and whistles. This surpasses Separate ViT-S by 8.1\%+ $\mu$AP / 6.7\%+ RP90. Even when compared to Separate ViT-B, CopyNCE still demonstrates a substantial advantage, which validates the effectiveness of CopyNCE in the standard setting. Furthermore, local ensembling could significantly enhances performance, resulting notable improvements of 6.7\%+ $\mu$AP / 5.7\%+ RP90 over Separate ViT-B under similar settings, even bringing CopyNCE ViT-S on par with Separate ViT-L. More surprisingly, resolution of $336\times336$ could boost CopyNCE to 88.7\% $\mu$AP / 83.9\% RP90, which exceeds D$^2$LV by 0.1\% $\mu$AP / 3.8\% RP90. Note that D$^2$LV ensembles 33 models, each of which has no less parameters than ViT-S.
        
        \noindent
        \textbf{Descriptor.} CopyNCE obtains 70.5\% $\mu$AP / 63.6\% RP90 without relying on additional data. When compared to other SOTAs with similar or larger models, CopyNCE outperforms S-square, Lykaap, and SSCD by 4.1\%+ $\mu$AP and surpasses Lykaap by 7.0\% RP90. It is noteworthy that although S-square utilized a smaller resolution, it benefits from additional data and a larger model. By applying post-processing, \textit{i.e.,} score normalization, CopyNCE can further enhance performance by 2.1\% $\mu$AP / 4.8\% RP90. In comparison with SSCD under similar setting, our method demonstrates a gain of 0.1\%+ $\mu$AP / 5.3\%+ RP90.

    \subsection{Ablation Studies and Parameter Analysis}
        To investigate contribution of components and settings, we perform ablation studies and parameter analysis. Results of descriptor and matcher are shown in Tab.~\ref{tab:descriptor_ablation_studies} and Tab.~\ref{tab:matching_ablation_studies}.
        \begin{table}[t]
            \small
            \begin{center}
            \begin{tabular}{ccccc}

            \hline
            
            \hline\hline
    
            Method & Parameter & $\mu$AP & Parameter & $\mu$AP \\
            \hline

            \multirow{10}{*}{CopyNCE} & \cellcolor[HTML]{ECF4FF}{ \textbf{default} } & \cellcolor[HTML]{ECF4FF}{\textbf{70.5}} & \cellcolor[HTML]{E8E8E8}{$ w_{\text{NCE}}=0 $ } & \cellcolor[HTML]{E8E8E8}{68.9}  \\
            \cline{2-5}

             & $ w_{\text{NCE}}=3 $ & 70.5 & $ w_{\text{NCE}}=8 $ & 69.9 \\
            \cline{2-5}

             & $ \gamma=0 $ & 67.9 & $ \gamma=0.5 $ & 69.7 \\
             & $ \gamma=1 $ & 70.0 & $ \gamma=2 $ & 70.4 \\
             & $ \gamma=3 $ & 70.5 & $ \gamma=+\infty $ & 70.1 \\ 
            \cline{2-5}

             & w/o NCE & 68.6 & layer=10 & 70.3 \\
            \cline{2-5}

             & \makecell{w/o GHNM \\$ w_{\text{NCE}}=0 $} & 57.7 & \makecell{w/o GHNM \\$ w_{\text{NCE}}=5 $} & 61.8 \\
            \cline{2-5}
            
             & \makecell{R50$\dag$ \\$ w_{\text{NCE}}=0 $} & 67.8 & \makecell{R50$\dag$ \\$ w_{\text{NCE}}=5 $} & 68.8 \\
            \hline
            
            \hline

            \hline

            FeatNN Cos & $k=1$ & 56.5 & $k=4$ & 48.1 \\
            FeatNN NCE & $k=1$ & 57.0 & $k=4$ & 42.6 \\

            \hline

            LocNN Cos & $k=1$ & 67.7 & $k=4$ & 67.2 \\
            LocNN NCE & $k=1$ & 64.7 & $k=4$ & 64.2 \\

            \hline
            
            Both Cos & $k=1$ & 68.5 & $k=4$ & 66.0 \\
            Both NCE & $k=1$ & 64.9  & $k=4$ & 64.3 \\

            \hline\hline
    
            \hline
            
            \end{tabular}
            \end{center}
            \vspace{-0.4cm}
            \caption{\textbf{Ablation studies} and \textbf{parameter analysis} on \textbf{descriptor}. \colorbox{copynce}{\textbf{default}} denotes the default parameters in Sec.~\ref{sec:implementation_details} and baseline is marked by \colorbox{baseline}{Grey}. \textbf{GHNM} means global hard negative mining, detailed in Supp. The lower part presents results of heuristic matching methods. \textbf{FeatNN}~\cite{li2021esvit, wang2022exploring, Yun2022PatchlevelRL} and \textbf{LocNN}~\cite{bardes2022vicregl} utilize feature and patch centroid $k$-NN respectively to match positives. And \textbf{Both} combines FeatNN and LocNN. Cos and NCE denote cosine loss and InfoNCE loss respectively. R50$\dag$ experiments are conducted under $448 \times 448$. More details are provided in Supp.} 

            \label{tab:descriptor_ablation_studies}
            \vspace{-0.4cm}
        \end{table}

        \noindent
        \textbf{Effectiveness of CopyNCE.} In descriptor scenario, CopyNCE results in an increase of 1.6\% $\mu$AP. Without global hard negative mining (GHNM) trick, improvements are even more pronounced, reaching 4.1\% $\mu$AP. To validate the generalization of CopyNCE, we follow the loss weight settings in SSCD~\cite{pizzi2022self} without any other parameter tuning and simply replace ViT-S with R50~\cite{He2016DeepRL}. CopyNCE still brings 1.0\% $\mu$AP boost. As for matcher, improvements of 12.6\% $\mu$AP are observed after adding CopyNCE on BCE, which collectively demonstrates the effectiveness of CopyNCE.

        \noindent
        \textbf{Effectiveness of Fine-grained Supervision.} To compare with heuristic local SSL methods, we replace PixTrace with FeatNN and LocNN. We explore various settings on descriptor and subsequently transfer some of the optimal to matcher. As mentioned in Sec.~\ref{sec:pixtrace}, both FeatNN and LocNN are subject to noise and encounter issues of false match and partial match. When comparing with FeatNN, LocNN could achieve more accurate supervision with additional location information. Consequently, LocNN demonstrates better performance and greater stability, as evidenced in results of both matcher and descriptor. When combining FeatNN and LocNN, we observe further improvements. Especially in matcher scenario, ``Both'' could significantly enhance baseline by 9.9\% $\mu$AP. We attribute this primarily to the fact that on the basis of LocNN, FeatNN is more stable and recalls more true positives to alleviate the partial match issue. To further mitigate partial match, we attempt to increase $k$ in $k$-NN selections. This modification not only increases the recall of positives, but also could lead to additional noise. Besides, it completely ignores various importance of positives. For descriptor, the results of $k=4$ fail to exhibit any advantages over $k=1$. Despite partial success of these heuristic methods, a consistent performance gap remains in comparison to PixTrace, which underscores the effectiveness of PixTrace in ICD.

        \noindent
        \textbf{NCE v.s. Cosine.} In ICD, assessing copy and non-copy relationships between regions is essential. NCE pulls positives closer while pushing negatives away, which facilitates the model to distinguish such relationships. Thus, NCE aligns more naturally with copy detection. ``w/o NCE'' in Tab.~\ref{tab:descriptor_ablation_studies} confirms this assertion. Moreover, ``w/o NCE'' fails to yield any improvement over ``Baseline''. This is mainly because solely using cosine loss will induce a trivial solution of the model, where all patch tokens are identical to each other. This prevents the descriptor from further fitting the target distribution provided by PixTrace. And this issue occurs with matcher as well and directly leads to its collapse with both FeatNN and LocNN. Another interesting finding is that for descriptor, NCE performs worse than Cos with noisy target distribution. 
        The reason is that when NCE faces false match, it incorrectly brings ``false'' positives closer and pushes ``true'' positives away, whereas Cos only pulls the ``false'' positives closer. Thus, NCE will impair performance of model more significantly than Cos.

        \noindent
        \textbf{Evaluation on $ w_{\text{NCE}} $.} For matcher, a smaller $ w_{\text{NCE}} $ leads to significant degradation, while increasing $ w_{\text{NCE}} $ to 3 saturates the performance. Meanwhile, both $ w_{\text{NCE}} =3 $ and $ w_{\text{NCE}} =5 $ obtain the best results of descriptor.

        \noindent
        \textbf{Evaluation on $\gamma$.} When $\gamma=0$, all positive patches get equal attention, which fails to reflect varying importance of each positive patch. Conversely, when $\gamma=+\infty$, the model only focuses on the positive patch with the largest overlapping area. Ultimately, the matcher demonstrates optimal performance when $\gamma=1$, which is proportional to overlapping ratio. In contrast to matcher, descriptor evaluates copy probability solely based on global feature similarity. The global features represent a fusion of all local features. For an edited copy patch in the qeury, in addition to fitting the target distribution provided by PixTrace, it is also essential to achieve high similarity with the positive patches in the reference to encourage agreement of query and reference global features. However, a more uniform allocation of attention is not conducive to achieving this objective. Thus, in consideration of this trade-off, $\gamma=3$ emerges as a more favorable choice.

        \noindent
        \textbf{Evaluation on matcher arch.} As shown in Tab.~\ref{tab:matching_ablation_studies}, a larger fusion block yields better performance in matcher, whereas a smaller one results in poorer outcomes. However, the computational cost of fusion layers is significantly greater than that of encoders, because it takes tokens from query and reference as input to perform cross attention. For trade-off, enc-8-fus-4 is a more applicable setting.

        \noindent
        \textbf{Evaluation on position of CopyNCE.} Descriptor is not sensitive to the position of CopyNCE. We cast CopyNCE at the 10th layer and obtain the similar results as those of the default settings. This further demonstrates the robustness of CopyNCE.
        
        \begin{table}[t]
            \small
            \begin{center}
            \begin{tabular}{ccccc}

            \hline
            
            \hline\hline
    
            Method & Parameter & $\mu$AP & Parameter & $\mu$AP \\
            \hline

            \multirow{7}{*}{CopyNCE} & \cellcolor[HTML]{ECF4FF}{ \textbf{default} } & \cellcolor[HTML]{ECF4FF}{\textbf{83.5}} & \cellcolor[HTML]{E8E8E8}{$ w_{\text{NCE}}=0 $ } & \cellcolor[HTML]{E8E8E8}{ 70.9 } \\
            \cline{2-5}

             & $ w_{\text{NCE}}=1 $ & 81.7 & $ w_{\text{NCE}}=5 $ & 83.5 \\
            \cline{2-5}

             & $ \gamma=0 $ & 82.5 & $ \gamma=0.5 $ & 82.6 \\
             & $ \gamma=1 $ & 83.5 & $ \gamma=2 $ & 82.9 \\
             & $ \gamma=3 $ & 83.0 & $ \gamma=+\infty $ & 82.6 \\
             \cline{2-5}
             
             & enc-6-fus-6 & 84.0 & enc-10-fus-2 & 79.4 \\
            \hline
            
            \hline
            
            \hline

            FeatNN  & Cos \; $k=1$ & Fail & NCE \; $k=1$ & Fail \\

            \hline

            LocNN & Cos \; $k=1$ & Fail & NCE \; $k=1$ & 78.7 \\

            \hline
            
            Both & Cos \; $k=1$ & Fail & NCE \; $k=1$ & 80.8 \\

            \hline\hline
    
            \hline
            
            \end{tabular}
            \end{center}
            \vspace{-0.4cm}
            \caption{\textbf{Ablation studies} and \textbf{parameter analysis} on \textbf{matcher}. ``enc-$m$-fus-$n$'' denotes encoder of $m$ layers and fusion of $n$ layers. \colorbox{copynce}{\textbf{default}} denotes the default parameters in Sec.~\ref{sec:implementation_details} and baseline is marked by \colorbox{baseline}{Grey}. The lower part presents results of heuristic matching methods.} 

            \label{tab:matching_ablation_studies}
            \vspace{-0.4cm}
        \end{table}

    \subsection{Visualization and Interpretability}
        \begin{figure*}[t]
            \begin{center}
                \includegraphics[width=0.8 \linewidth]{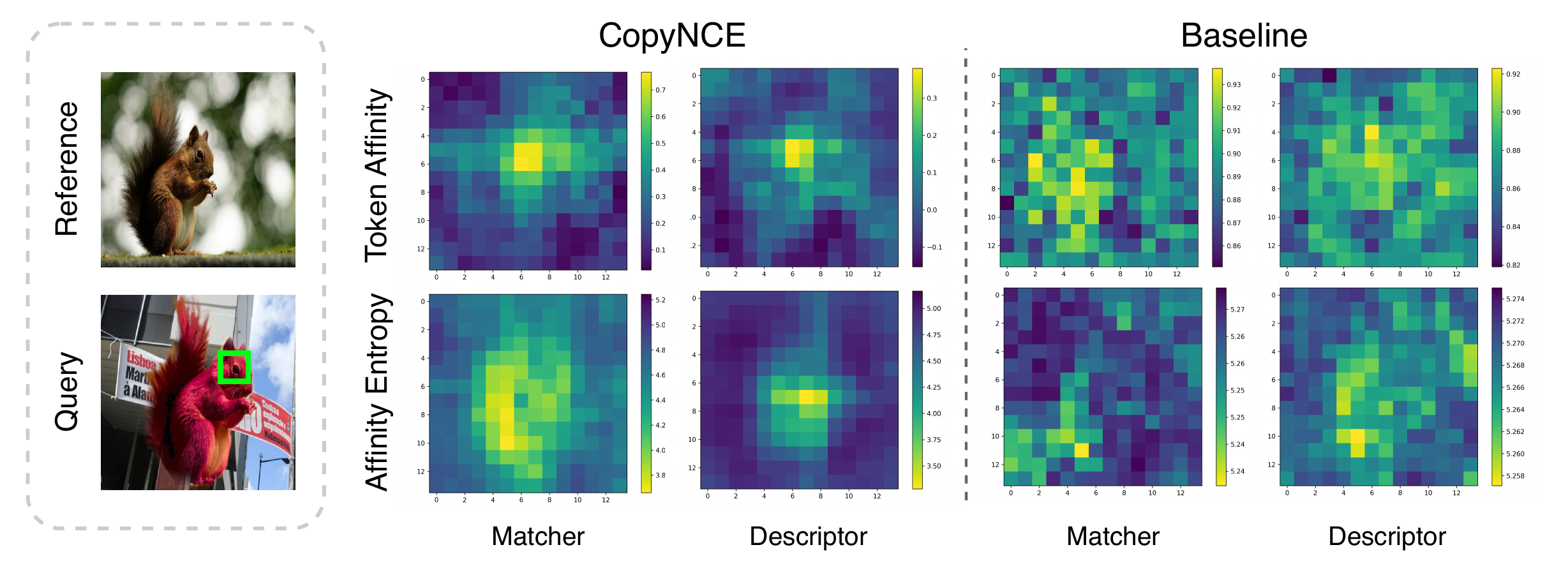}
            \end{center}
            \vspace{-0.4cm}
            \caption{\textbf{Visualization of token affinity and affinity entropy.} The left part displays reference and query images. \textcolor[RGB]{0, 255, 0}{Green box} is the probe patch to draw token affinity heatmaps. The subsequent section contains token affinity heatmaps and affinity entropy of different models. }
            \label{fig:visualization}
            \vspace{-0.2cm}
        \end{figure*}

    According to the motivation of CopyNCE, if a token from copy region acts as a probe, its corresponding copied patch in reference should be prominently emphasized in affinity heatmap. To visualize this, we take instances in Fig.~\ref{fig:visualization}, where query copies upon the squirrel in reference. We select the token representing eye of squirrel from query as the probe. By applying CopyNCE, tokens near the copy location are highlighted in affinity matrices of both matcher and descriptor. In contrast, the affinity heatmap appears chaotic with baseline, failing to reveal effective copy correspondences.

    To further demonstrate the ability to capture copy regions, we introduce affinity entropy $\mathscr{E}$:
    \begin{equation}
        \mathscr{E}_i = -\sum_j p_{ij} \log p_{ij}, \; p_{ij} = \frac{\exp( \cos( z_i^q, z_j^r ) / \tau)}{ \sum_k \exp( \cos( z_i^q, z_k^r ) / \tau )}.
        \label{equ:affinity_entropy}
    \end{equation}
    $\mathscr{E}_i$ measures the affinity entropy of query token $z_i^q$. For the tokens of a copy patch, affinity heatmap should accurately focus on tokens of the copied patches, rendering a smaller entropy. After calculating affinity entropy of all query tokens, affinity entropy heatmaps are visualized in Fig.~\ref{fig:visualization}. For ``CopyNCE'' group, tokens covering the squirrel are distinctly brighter than the surrounding, which forms a clear outline. This indicates that the entire squirrel is identified as edited copy region by model. While ``Baseline'' group completely fails to reveal any evident outline of the squirrel. This comparison suggests that CopyNCE not only improves performance but also enhances interpretability.

    \subsection{More Discussion}
        \subsubsection{Further comparison with LocNN}
            In cases where images are augmented by simple transformations, \textit{e.g.} cropping, scaling and flipping, LocNN actually could use patch centroids along with four vertices of patches to determine whether there is an overlap between patches. However, when faced with more complex edits, such as rotation, stretching, or image matting, which might lead to deformed and irregular copy regions, LocNN becomes nearly ineffective. And all these complex transforms are quite common in image copy behavior. To conclude, LocNN is a reasonable approximation of PixTrace, particularly in tasks that emphasize semantic features. However, in the context of copy detection, PixTrace is essential.

            \begin{table}[t]
                \small
                \begin{center}
                \begin{tabular}{ccc|ccc}
            
                \hline
                
                \hline\hline
            
                \multicolumn{3}{c|}{Descriptor Track} & \multicolumn{3}{c}{Matching Track} \\
                \hline
            
                Team & $\mu$AP & RP90 & Team & $\mu$AP & RP90 \\
                \hline
                
                \cellcolor[HTML]{ECF4FF}{\textbf{CopyNCE\dag}} & \cellcolor[HTML]{ECF4FF}{\textbf{65.8}} & \cellcolor[HTML]{ECF4FF}{\textbf{61.0}} & \cellcolor[HTML]{ECF4FF}{\textbf{CopyNCE\dag}} & \cellcolor[HTML]{ECF4FF}{\textbf{85.6}} & \cellcolor[HTML]{ECF4FF}{\textbf{80.0}} \\
            
                lyakaap\dag & 63.5 & 55.4 & \cellcolor[HTML]{ECF4FF}{\textbf{CopyNCE}} & \cellcolor[HTML]{ECF4FF}{\textbf{84.6}} & \cellcolor[HTML]{ECF4FF}{\textbf{78.2}} \\
    
                \cellcolor[HTML]{ECF4FF}{\textbf{CopyNCE}} & \cellcolor[HTML]{ECF4FF}{\textbf{60.9}} & \cellcolor[HTML]{ECF4FF}{\textbf{56.7}} & VisionForce & 83.3 & 73.1 \\
                
                S-square & 59.1 & 50.9 & separate\dag & 82.9 & 79.2 \\
            
                visionForce & 57.9 & 48.9 & imgFp\dag & 76.8 & 67.2 \\
                
                \hline\hline
            
                \hline
                
                \end{tabular}
                \end{center}
                \vspace{-0.4cm}
                \caption{\textbf{Leaderboard of DISC21 Phase 2.} \dag \; denotes the results achieved after finetuning on \texttt{dev set part I}. Note that finetuning is allowed by official rules~\cite{pmlr-v176-papakipos22a}.} 
            
                \label{tab:leaderboard_of_phase_2}
                \vspace{-0.4cm}
            \end{table}

        \subsubsection{Performance on DISC21 phase 2}
            
            To thoroughly evaluate CopyNCE, we adhered to rules of DISC21 to simulate test pipeline of phase 2, which contains 50k query images with aggressive editing than test of phase 1. Tab.~\ref{tab:leaderboard_of_phase_2} shows that CopyNCE achieves significant improvements over all open-source methods on both tasks with or without finetuning.

        \subsubsection{Performance on NDEC}
            \begin{table}[t]
                \small
                \begin{center}
                \begin{tabular}{cccc}
        
                \hline
                
                \hline\hline
                
                Method & Model Arch. & $\mu$AP & RP90 \\
                \hline
        
                \cellcolor[HTML]{ECF4FF}{\textbf{CopyNCE}} & \cellcolor[HTML]{ECF4FF}{\textbf{ViT-B+ViT-S}} & \cellcolor[HTML]{ECF4FF}{\textbf{72.5}} & \cellcolor[HTML]{ECF4FF}{\textbf{36.8}}\\ 
        
                \hline
        
                Strong ASL & Multi & 64.1 & - \\
                \hline
        
                D$^2$LV ASL & Multi & 61.3 & - \\
        
                \hline\hline
        
                \hline
                
                \end{tabular}
                \end{center}
                \vspace{-0.4cm}
                \caption{\textbf{Results on NDEC.} \textbf{Multi} in ``D$^2$LV ASL'' and ``Strong ASL'' stands for 11$\times$R50, 11$\times$R152 and 11$\times$R50IBN. } 
        
                \label{tab:ndec}
                \vspace{-0.4cm}
            \end{table}
            To further investigate the potential of CopyNCE, we directly adopt CopyNCE on NDEC without finetuning on \texttt{dev set part I} or its extended \texttt{training set}. We conduct inference pipeline of DISC21 Phase 2 on NDEC, CopyNCE achieves 72.5\% $\mu$AP, surpassing Strong ASL by 8.4\% $\mu$AP. We attribute the impressive boost to natural local modeling of matcher. The primary challenge in handling hard negatives lies in their subtle distinctions at the fine-grained level. The attention mechanism of matcher facilitates the exchange of regional information between query and reference and enables detailed comparison of patches. Despite ensembling in other methods, each of these models serves as a descriptor, making it impossible to conduct such a detailed comparison.

\section{Conclusion}
    \label{sec:conclusion}
    In this work, motivated by \textit{inherent traceability of edited copy pixels}, we developed PixTrace to track coordinates of pixels after various copy edits. To provide pixel-level supervision, we proposed CopyNCE. It takes the copy area proportion between patches as prior target distribution to regularize affinity of patches. Finally, CopyNCE series achieved SOTA results of 88.7\% $\mu$AP / 83.9\% RP90 for matcher and 72.6\% $\mu$AP / 68.4\% RP90 for descriptor. Extensive experiments also validated its impressive generalization, efficiency and interpretability.

{
    \small
    \bibliographystyle{ieeenat_fullname}
    \bibliography{main}
}

\clearpage
\setcounter{page}{1}

\maketitlesupplementary
\setcounter{equation}{0}
\setcounter{section}{0}
\renewcommand{\theequation}{A.\arabic{equation}}
\renewcommand{\thesection}{A.\arabic{section}}

\section{More about DISC21 Data}
    The DISC21 competition~\cite{douze20212021} comprises four datasets: \texttt{training set}, \texttt{reference set}, \texttt{dev set} (which is evenly split into two parts: \texttt{dev set part I} and \texttt{dev set part II}) and \texttt{test set}. All data in \texttt{training set} is unlabeled. Both \texttt{dev set} and the \texttt{test set} function as queries, sharing \texttt{reference set}. According to the rules outlined on the DISC21 official website\footnotemark[1], the competition is divided into two phases. In the first phase, \texttt{dev set part I} and its labels are made public, serving as a validation set for the participants. Meanwhile, \texttt{dev set part II} is also made public, but its labels remain concealed for the purpose of ranking participants' performance in the first phase. Both data and labels in \texttt{test set} will not be disclosed during this phase. In the second phase, images in \texttt{test set} will be released for the final rankings. After the competition, labels for both \texttt{dev set part II} and \texttt{test set} will be accessible. For clarity, please refer to Tab.~\ref{tab:disc21_datasets}. It is noteworthy that the competition rules allow the fine-tuning on \texttt{dev set part I}. Since times of evaluating on  \texttt{dev set part II} during the first phase are no limited, while times of evaluation on \texttt{test set} is restricted, there are more reported results on \texttt{dev set part II} from public papers than those on \texttt{test set}. Therefore, Tab~\ref{tab:comparison_with_sotas} of this paper utilizes results on \texttt{dev set part II}, while the simulated results on \texttt{test set} are presented in Tab.~\ref{tab:leaderboard_of_phase_2}.

    \footnotetext[1]{https://www.drivendata.org/competitions/group/image-similarity-challenge}
    
    \begin{table*}[h]
        \begin{center}
        \begin{tabular}{cc|cccccccc}
        \hline
                
        \hline\hline
        &  & \multirow{2}{*}{Training Set} & \multirow{2}{*}{Reference Set} & \multicolumn{2}{c}{Dev Set Part I} & \multicolumn{2}{c}{Dev Set Part II} & \multicolumn{2}{c}{Test Set} \\

        &  &  &  & Data & Labels & Data & Labels & Data & Labels \\
        \hline

        \multicolumn{2}{c|}{Scale} & 1,000,000 & 1,000,000 & \multicolumn{2}{c}{25,000} & \multicolumn{2}{c}{25,000} &  \multicolumn{2}{c}{50,000} \\

        \multirow{3}{*}{Accessible} & Phase 1 & \Checkmark & \Checkmark & \Checkmark & \Checkmark & \Checkmark & \XSolidBrush & \XSolidBrush & \XSolidBrush \\

        & Phase 2 & \Checkmark & \Checkmark & \Checkmark & \Checkmark & \Checkmark & \XSolidBrush & \Checkmark & \XSolidBrush \\

        & After & \Checkmark & \Checkmark & \Checkmark & \Checkmark & \Checkmark & \Checkmark & \Checkmark & \Checkmark \\
                
        \hline\hline
        
        \hline
        \end{tabular}
        \end{center}
        \caption{\textbf{Details of DISC21 datasets.} In ``Accessible'' part, \Checkmark denotes that corresponding data is available during that period of time, and \XSolidBrush denotes unavailability.} 
        
        \label{tab:disc21_datasets}
    \end{table*}

\section{More about Evaluation Protocols}
\label{sec:more_about_evaluation_protocols}
    \begin{figure}[ht]
        \begin{center}
            \includegraphics[width=1.0 \linewidth]{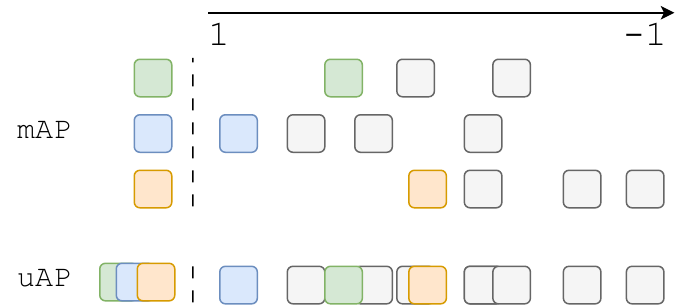}
        \end{center}
        \caption{\textbf{mAP \textit{v.s.} $\mathbf{\mu}$AP.} In this case, mAP is 100\%, because all positive samples are ranked at the first place in the view of each query. While no threshold could be set to achieve that perfect performance. In $\mu$AP, results are concatenated together to calculate precision of each threshold. Thus, $\mu$AP could better reflect the real performance of model. Mathematically, mAP also serves as an upper bound of $\mu$AP.}
        \label{fig:mAP_vs_uAP}
    \end{figure}

    The formulas for calculating mAP (mean Average Precision) and $\mu$AP (unified Average Precision) are as follows: 
    \begin{equation}
        \text{mAP} = \frac{1}{N} \sum_{i=1}^{N} \text{AP}( \{ s_{ij} | j=1, 2, \dots \} ),
    \end{equation}
    \begin{equation}
        \mu\text{AP} = \text{AP}(\{ s_{ij} \} ),
    \end{equation}
    where $s_{ij}$ denotes similarity score between query i and reference j. mAP averages Average Precision (AP) of each sample, while $\mu$AP concatenates similarity scores of all image pairs to compute AP. Consequently, under mAP formulation, the distribution of similarity scores for each query may vary significantly. While in practical scenarios, a fixed threshold is typically used to filter edited copy pairs. In cases where the distribution of similarity scores varies greatly, mAP may not accurately reflect the performance of model, as the case in Fig.~\ref{fig:mAP_vs_uAP}. In contrast, $\mu$AP computes AP based on all aggregated scores, which better simulates real-world scenario.

\section{More about Post-Processing}
    \subsection{Score Normalization}
        Score normalization (SN) is a widely used technique in data mining. SN maps data from different distributions to a common distribution. If the distributions of similarity score vary among queries, a global threshold is difficult to be set. Then SN could effectively address this problem and improve $\mu$AP significantly. We refer to SN method in SSCD~\cite{pizzi2022self}, which is formulated as follows:
        \begin{equation}
            s_{ij} = \cos(z_i, z_j) - \alpha \frac{1}{k_{\text{end}} - k_{\text{start}} + 1} \sum_{k=k_{\text{start}}}^{k_{\text{end}}} \cos(z_i, z_{k\text{-NN}}) ,
        \end{equation}
        where $s_{ij}$ is normalized score of query $x_i$ and reference $x_j$, $z_i$ and $z_j$ are their features extracted by descriptor. $k$-NN is the top-$k$ nearest neighbor of query $x_i$ in auxiliary data. Here we simply employ \texttt{training set} as auxiliary data and set $k_{\text{start}}=0$, $k_{\text{end}}=9$, $\alpha=1$, which means that takes the closest neighbor in \texttt{training set} as bias.

    \subsection{Feature Stretching}
        \label{sec:feature_stretching}
        Feature stretching a technology to stabilize scores across different queries. It is introduced by Wang \textit{et al.} in BoT~\cite{Wang2021BagOT}. In short, query feature $z_q$ is stretched by formula below:
        \begin{equation}
            \bar{z_q} = \beta \frac{\sum_{i=1}^k s_i}{k} z_q,
        \end{equation}
        where $\beta$ is stretching coefficient, $\{ s_1, s_2, \dots, s_k \}$ denotes the similarity scores (inner product) of $k$-NN images of query in the auxiliary dataset. After all query features are stretched, euclidean distance is utilized to ranking similarity between query and reference. In this paper, we follow the settings in BoT~\cite{Wang2021BagOT} that $\beta = 2.5$, $k=5$ and let training set as the auxiliary set.

\section{Experiments on more datasets}

    \begin{table}[t]
        \footnotesize
        \begin{center}
        \begin{tabular}{c|cc|c|cc}
        
        Method & $\mu$AP & R@1 & Method & $\mu$AP & R@1 \\
        \hline
        SSCD~\cite{pizzi2022self} & 14.22 & 20.24 & \cellcolor[HTML]{ECF4FF}{\textbf{ViT-S}} & \cellcolor[HTML]{ECF4FF}{\textbf{27.07}} & \cellcolor[HTML]{ECF4FF}{\textbf{34.68}} \\
        
        S-square~\cite{Papadakis2021ProducingAE} & 14.51 & 21.05 & \cellcolor[HTML]{ECF4FF}{\textbf{ViT-B}} & \cellcolor[HTML]{ECF4FF}{\textbf{31.66}} & \cellcolor[HTML]{ECF4FF}{\textbf{37.78}} \\
    
        Lyakaap~\cite{yokoo2021contrastive} & 13.80 & 18.02 & \cellcolor[HTML]{ECF4FF}{\uline{ViT-S}}\dag & \cellcolor[HTML]{ECF4FF}{\uline{25.38}} & \cellcolor[HTML]{ECF4FF}{\uline{31.57}} \\
    
        AnyPat. Base.~\cite{wang2024anypattern} & 16.18 & 20.54 & \cellcolor[HTML]{ECF4FF}{\uline{ViT-B}}\dag & \cellcolor[HTML]{ECF4FF}{\uline{28.05}} & \cellcolor[HTML]{ECF4FF}{\uline{34.36}}
        
        \end{tabular}
        \end{center}
        \vspace{-0.3cm}
        \caption{\textbf{Results on AnyPattern.} All methods are evaluated with ``SmallPattern'' protocol . ``AnyPat. Base.'' denotes Baseline in AnyPattern. CopyNCE results are marked in \colorbox{copynce}{blue} and \dag \; means results achieved with augmentations that aligned with Lyakaap. } 
    
        \label{tab:any_pattern}
        \vspace{-0.3cm}
    \end{table}

    \begin{table}[t]
        \small
        \begin{center}
        \begin{tabular}{c|ccc}
        
        & SSCD SN~\cite{pizzi2022self} & \textbf{\cellcolor[HTML]{ECF4FF}{ViT-S} SN} & \textbf{\cellcolor[HTML]{ECF4FF}{ViT-B} SN} \\
        \hline
        Descriptor $\mu$AP & 64.99 & \cellcolor[HTML]{ECF4FF}{70.59} & \textbf{\cellcolor[HTML]{ECF4FF}{71.57}} \\ 
        Matching $\mu$AP & 46.92 & \textbf{\cellcolor[HTML]{ECF4FF}{51.32}} & \cellcolor[HTML]{ECF4FF}{50.05}\\
        
        \end{tabular}
        \end{center}
        \vspace{-0.3cm}
        \caption{\textbf{Results on VSC2022.} Results are produced by official baseline implementation of VSC2022 on its training set.} 
    
        \label{tab:vsc2022}
        \vspace{-0.3cm}
    \end{table}

    \textbf{AnyPattern~\cite{wang2024anypattern}.} To evaluate the generalization of CopyNCE to unseen copy edits, we conducted experiments on AnyPattern~\cite{wang2024anypattern} test set. Note that models are only trained on \texttt{training set} of DISC21, without any finetuning on \texttt{dev set part I}. AnyPattern encompasses rare and aggressive copy edits that are completely disjoint from the augmentation pipeline employed during CopyNCE training. As reported in Tab.~\ref{tab:any_pattern}, under the ``SmallPattern'' setting, CopyNCE with ViT-S achieves 27.07\% $\mu$AP and 34.68\% R@1, whereas the ViT-B variant attains 31.66\% $\mu$AP and 37.78\% R@1. Although our original augmentation is more aggressive than competing approaches, we further aligned it with the augmentation settings of Lyakaap~\cite{yokoo2021contrastive} and retrained CopyNCE from scratch. Even under this reduced augmentation, CopyNCE still achieves 25.38\% $\mu$AP / 31.57\% R@1 (ViT-S) and 28.05\% $\mu$AP / 34.36\% R@1 (ViT-B). Despite the expected performance drop due to weaker augmentation, these results exceed the best competing method by at least 9.20\%+ $\mu$AP and 10.52\%+ R@1. This empirical evidence corroborates that CopyNCE possesses superior generalization to previously unseen infringement patterns.
     
    \textbf{VSC2022~\cite{pizzi20242023}.} To further validate CopyNCE in the context of video copy detection, we evaluate the model trained on DISC21 by following the official baseline\footnotemark[1] of VSC2022~\cite{pizzi20242023} and testing it on the VSC2022 training set. Under score normalization, CopyNCE surpasses SSCD by 5.6\%+ Descriptor $\mu$AP and 3.1\%+ Matching $\mu$AP. These results indicate that CopyNCE, when employed as a feature extractor for video copy detection, still delivers impressive performance.

    \footnotetext[1]{https://github.com/facebookresearch/vsc2022/blob/main/docs/baseline.md}

\section{Scaling Performance}
    \begin{table}[t]
        \small
        \setlength{\tabcolsep}{1.6mm}
        \begin{center}
        \begin{tabular}{cccccc}

        \hline
        
        \hline\hline

        \multirow{2}{*}{\makecell{Test\\Set}} & \multirow{2}{*}{Model} & \multirow{2}{*}{Resolution} & \multicolumn{3}{c}{Metrics} \\
        \cline{4-6}

         &  &  & mAP & $\mu$AP & RP90 \\
        \hline

        \texttt{Dev II} & ViT-S & \multirow{2}{*}{$224 \times 224$} & 90.6 & 83.5 & 75.4 \\
        \texttt{Dev II} & ViT-B &  & 90.5 & 83.5 & 76.7 \\
        \cline{3-6}
        \texttt{Dev II} & ViT-S & \multirow{2}{*}{$336 \times 336$} & 91.3 & 85.8 & 79.9 \\
        \texttt{Dev II} & ViT-B &  & 91.3 & 85.7 & 80.1 \\
        
        \hline\hline

        \hline
        
        \end{tabular}
        \end{center}
        \vspace{-0.4cm}
        \caption{\textbf{Scaling of matcher.} All results are achieved without finetuning on \texttt{dev set part I}.} 

        \label{tab:scaling_of_matching_model}
        \vspace{-0.4cm}
    \end{table}

    \begin{table}[t]
        \small
        \begin{center}
        \setlength{\tabcolsep}{1.6mm}
        \begin{tabular}{cccccc}

        \hline
        
        \hline\hline

        \multirow{2}{*}{\makecell{Test\\Set}} & \multirow{2}{*}{Model} & \multirow{2}{*}{Resolution} & \multicolumn{3}{c}{Metrics} \\
        \cline{4-6}

         &  &  & mAP & $\mu$AP & RP90 \\
        \hline

        \texttt{Dev II} & ViT-S  & \multirow{2}{*}{$224 \times 224$} & 76.5 & 70.5 & 63.6 \\
        \texttt{Dev II} & ViT-B &  & 77.9 & 72.3 & 65.2 \\
        \cline{3-6}
        \texttt{Dev II} & ViT-S & \multirow{2}{*}{$336 \times 336$} & 75.0 & 69.8 & 63.9 \\
        \texttt{Dev II} & ViT-B &  & 76.2 & 71.3 & 66.0 \\
        
        \hline\hline

        \hline
        
        \end{tabular}
        \end{center}
        \vspace{-0.4cm}
        \caption{\textbf{Scaling of descriptor.} All results are achieved without finetuning on \texttt{dev set part I}.} 

        \label{tab:scaling_of_descriptor}
        \vspace{-0.4cm}
    \end{table}

    Scaling performance matters in cases of subtle and complex edits. In this section, we explored how resolution and model size affect performance. For matcher, increasing resolution leads to improvements of 2.2\%+ $\mu$AP / 3.4\%+ RP90. However, unlike results of ``Separate'' shown in Tab.~\ref{tab:comparison_with_sotas}, enlarging model size for CopyNCE fails to boost performance. We hypothesize that matcher allows interaction between query and reference tokens, while CopyNCE provides direct supervision for interaction. Such guidance is sufficient enough for either ViT-S or larger ViT-B to tell whether query edits upon reference. And this will lead to performance saturation in terms of model size. This could be another evidence of effectiveness of CopyNCE.
    
    For descriptor, scaling model size brings lifts of 1.5\%+ $\mu$AP / 1.6\%+ RP90 in Tab.~\ref{tab:scaling_of_descriptor}, which is consistent with expectations. However, increasing resolution has negative effects. We believe the primary reason is that training descriptor requires a larger batch size. However, increasing the resolution significantly raises the memory consumption for ViT. To accommodate training at a higher resolution, it is necessary to reduce the batch size, which leads to the decline in descriptor performance. Another potential reason will be discussed in Sec.~\ref{sec:resnet_discussion}.

\section{More about Implementation Details}
    
    \label{sec:more_about_implementation_details}
    \begin{table*}[h]
        \begin{center}

        \begin{tabular}{cc|c|c}
        \hline
                
        \hline\hline

        \multicolumn{2}{c|}{\multirow{2}{*}{Settings}} & \multicolumn{2}{c}{Parameters} \\
        \cline{3-4}
        
        &  & \makebox[0.3\textwidth][c]{Matching} & \makebox[0.3\textwidth][c]{Descriptor} \\
        \hline

        & Resolution & \multicolumn{2}{c}{ $224 \times 224$ / $336 \times 336$ } \\
        \cline{2-4}
        
        \multirow{4}{*}{Input} & Augmentation & \multicolumn{2}{c}{\makecell{color jitter, random grayscale, random blur, overlay text, \\ overlay emoji, random flipping, affine, perspective, random resized-crop, \\ overlay image, random erasing, etc.}} \\
        \cline{2-4}

        & Positive Rate & 0.3 & - \\
        \cline{2-4}

        & \makecell{Hard Negative\\Mining} & p=0.5, \textit{k}-NN (\textit{k}=128) & p=1.0, \textit{k}-NN (\textit{k}=8) \\
        \hline

        \multirow{2}{*}{Model} & Arch & \makecell{ViT-S/16 \; ViT-B/16~\cite{dosovitskiy2021an} \\8 Encoder Layers + 4 Fusion Layers } & ViT-S/16 \; ViT-B/16~\cite{dosovitskiy2021an} \\
        \cline{2-4}

        & Linear Head & - & 512 dims \\
        \hline
        
        \multirow{10}{*}{Optim} & Pretraining & DINO~\cite{Caron2021EmergingPI} \; ViT-S/16 \; ViT-B/16\footnotemark[2] & DINO~\cite{Caron2021EmergingPI} \; ViT-S/16 \; ViT-B/16\footnotemark[2] \\
        \cline{2-4}

        & Batch Size & 8 GPUs $\times$ 32 & 8 GPUs $\times$ 96 \\
        \cline{2-4} 

        & Epoch & 30 ($\text{base lr}=1e-3$) + 30 ($\text{base lr}=2e-4$) & 30 ($\text{base lr}=6e-4$) \\
        \cline{2-4}

        & Optimizer & \multicolumn{2}{c}{AdamW ($\beta$=[0.9, 0.999 ])} \\
        \cline{2-4}
        
        & Weight Decay & \multicolumn{2}{c}{0.04} \\
        \cline{2-4} 
        
        & Learning Rate & $ 0.001 \times \sqrt{\text{bs} / 1024} $ & $ 0.0006 \times \sqrt{\text{bs} / 1024} $ \\
        \cline{2-4}

        & Scheduler & \multicolumn{2}{c}{Cosine Scheduler} \\
        \cline{2-4}

        & Min lr & \multicolumn{2}{c}{$2.0e-06$} \\
        \cline{2-4}
        
        & Warmup Epoch & \multicolumn{2}{c}{1} \\
        \cline{2-4}

        & Clip Grad & \multicolumn{2}{c}{3.0} \\
        \hline

        \multirow{2}{*}{Loss} & Baseline Loss & $\text{BCELoss} \times 1$ & $\text{InfoNCE} \times 1 + \text{KeLeoLoss} \times 5$ \\
        \cline{2-4}

        & CopyNCE & \multicolumn{2}{c}{Refer to Sec.~\ref{sec:experiments}} \\
        
        \hline\hline
        
        \hline
        
        \end{tabular}
        \end{center}
        
        \caption{\textbf{All detailed settings in our training pipeline.} For the aspects of data augmentation that have been omitted, please refer to our code.} 
        
        \label{tab:all_training_detailed_settings}
    \end{table*}
    \footnotetext[2]{https://github.com/facebookresearch/dino}

    \begin{table*}[h]
        \begin{center}

        \begin{tabular}{cc|c|c}
        \hline
                
        \hline\hline

        \multicolumn{2}{c|}{\multirow{2}{*}{Settings}} & \multicolumn{2}{c}{Parameters} \\
        \cline{3-4}
        
        &  & \makebox[0.3\textwidth][c]{Matching} & \makebox[0.3\textwidth][c]{Descriptor} \\
        \hline

        & Resolution & \multicolumn{2}{c}{ $224 \times 224$ / $336 \times 336$ } \\
        \cline{2-4}
        
        \multirow{4}{*}{Input} & Augmentation & \multicolumn{2}{c}{resize} \\
        \cline{2-4}

        & Positive Rate & 0.3 & - \\
        \cline{2-4}

        & \makecell{Hard Negative\\Mining} & Off & Off \\
        \hline

        Model & Linear Head & - & 256 dims \\
        \hline
        
        \multirow{4}{*}{Optim} & Pretraining & CopyNCE & CopyNCE \\
        \cline{2-4}

        & Batch Size & 8 GPU $\times$ 32 & 1 GPU $\times$ 96 \\
        \cline{2-4} 

        & Epoch & 20 ($\text{base lr}=2 \times 10^{-4}$) & 30 ($\text{base lr}=1 \times 10^{-4}$) \\
        \cline{2-4}
        
        & Learning Rate & $ 2 \times 10^{-4} \times \sqrt{\text{bs} / 1024} $ & $ 1 \times 10^{-4} \times \sqrt{\text{bs} / 1024} $ \\
        \hline

        \multirow{2}{*}{Loss} & Baseline Loss & $\text{BCELoss} \times 1$ & $\text{InfoNCE} \times 1 + \text{KeLeoLoss} \times 5$ \\
        \cline{2-4}

        & CopyNCE & \multicolumn{2}{c}{Off} \\
        
        \hline\hline
        
        \hline
        
        \end{tabular}
        \end{center}
        
        \caption{\textbf{All detailed settings in our finetune pipeline.} If certain settings are not listed in this table, they are set to be the same settings used in training by default.} 
        
        \label{tab:all_finetune_detailed_settings}
    \end{table*}

    All our training and finetune implementation details are listed in Tab.~\ref{tab:all_training_detailed_settings} and Tab.~\ref{tab:all_finetune_detailed_settings}. Note that due to numerical stability, we re-implemented the average and one-hot mode of CopyNCE in order to conduct experiments of $ \gamma = 0 $ and $ \gamma = +\infty $. And in finetuning, no pixel mappings are available due to human-made edits. Thus, only baseline loss is utilized for descriptor and matcher.

\section{More about Model Arch}
    \begin{table}[h]
        \footnotesize
        \setlength{\tabcolsep}{1.6mm}
        \begin{center}
        \begin{tabular}{c|cccc}

        & Encoder & Fusion & Extensive & Task \\

        \hline

        Descriptor & Att. Block & Cosine & 1 \textit{v.s.} N & Coarse Retrieval \\

        Matcher & Att. Block & Att. Block & 1 \textit{v.s.} 1 & Fine Matching \\
        
        \end{tabular}
        \end{center}
        \caption{\textbf{Property of descriptor and matcher.} ``Att. Block'' denotes Attention block for short. Descriptor extends well because it retrieves images through vector similarity and can yield N results with a single matrix multiplication. While matcher has to perform classification pair by pair.} 
        
        \label{tab:model_property}
    \end{table}
    \begin{figure}[ht]
    \begin{center}
        \includegraphics[width=1.0 \linewidth]{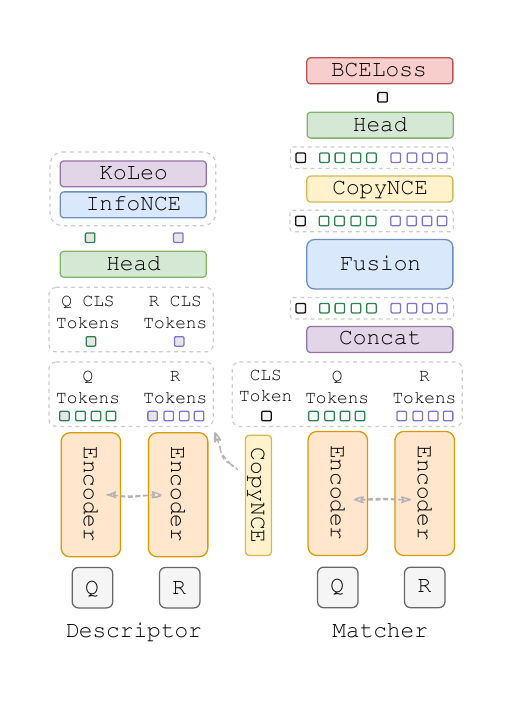}
    \end{center}
    \caption{\textbf{Detailed architecture of matcher and descriptor.}}
    \label{fig:detailed_architecture}
    \end{figure}

    The architectures of both matcher and descriptor are illustrated in Fig.~\ref{fig:detailed_architecture}. Matcher comprises an encoder and a fusion module, both constructed from attention blocks of same architecture. Matcher takes query and reference images as input, encoding each through encoder to obtain patch tokens. Subsequently, a learnable \texttt{[CLS]} token is concatenated with the tokens from both query and reference. And then they are passed to the fusion module for information interaction. Finally, CopyNCE supervises fused tokens, while binary cross-entropy (BCE) loss optimizes \texttt{[CLS]} token through a fully connected head. To enhance efficiency and reduce computational load, matcher is constructed based on a default ViT-S, which has 12 layers of attention blocks. The first eight layers form the encoder, while the last four layers constitute the fusion.
    
    In contrast, descriptor is solely based on ViT. Upon receiving query and reference, descriptor extracts their \texttt{[CLS]} tokens and patch tokens. In baseline scenario, \texttt{[CLS]} tokens of both images are trained with InfoNCE~\cite{Oord2018RepresentationLW} and KoLeo~\cite{sablayrolles2018spreading} loss. Within our framework, CopyNCE regularizes patch tokens in the last layer. Similarly, descriptor defaults to using ViT-S for simplification.

    Additionally, the property of both matcher and descriptor are listed in Tab.~\ref{tab:model_property}.
    
\section{More about Tricks}

    \subsection{Global Hard Negative Mining}
        \begin{figure}[ht]
            \begin{center}
                \includegraphics[width=1.0 \linewidth]{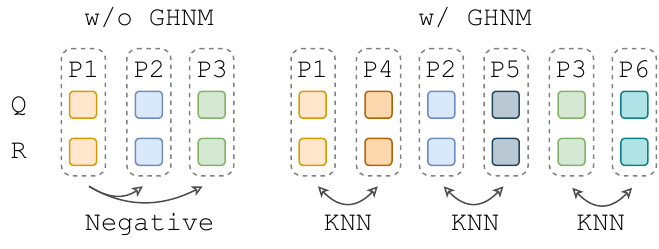}
            \end{center}
            \caption{\textbf{Demonstration of GHNM.}}
            \label{fig:ghnm}
        \end{figure}
    
        \begin{table}[t]
            \begin{center}
            \begin{tabular}{ccccc}
    
            \hline
            
            \hline\hline
    
            \multirow{2}{*}{Method} & \multirow{2}{*}{Parameter} & \multicolumn{3}{c}{Metrics} \\
            \cline{3-5}
    
            &  & mAP & $\mu$AP & RP90 \\
            \hline
    
            Baseline &  & 76.4 & 68.9 & 60.6 \\
            \hline
            
            Baseline & w/o GHNM & 77.1 & 57.7 & 30.0 \\
    
            \hline\hline
    
            \hline
            
            \end{tabular}
            \end{center}
            \caption{\textbf{Ablation studies of GHNM} on \textbf{descriptor}.} 
    
            \label{tab:descriptor_GHNM_ablation_studies}
        \end{table}

        \label{alg:ghnm}
        During training of descriptor, although hard negative mining (HNM) is performed in both KoLeo and InfoNCE, the mining process is constrained within batch, which is limited by batch size. Thus, for a training set consisting of millions of images, the probability of identifying hard samples in $k$-NNs is minimal. To address this issue, we utilize global hard negative mining (GHNM) as Algo.~\ref{alg:ghnm}. This method ensures that every sample in the mini-batch can at least find a $k$-NN-level negative sample, significantly improving the lower bound for hard negative mining. We conduct ablation studies on GHNM and the results are reported in Tab.~\ref{tab:descriptor_GHNM_ablation_studies}. Compared to baseline w/o GHNM, which employs standard hard negative mining (HNM) within the batch, GHNM leads to substantial 7.1\% $\mu$AP / 30.6\% RP90 enhancement for descriptor, clearly demonstrating the necessity of GHNM.
        \begin{algorithm}[h]
        \caption{Global Hard Negative Mining (GHNM)}
        \small
        \hspace*{0.02in}{\bf Input:}
        Batch Size $ N $. \\
        \hspace*{0.02in}{\bf Input:}
        \texttt{KNN} constructed by DINO~\cite{Caron2021EmergingPI} features. \\
        \hspace*{0.02in}{\bf Output:}
        Mini-batch $\{ [x, x']_i | i=1, \dots, N \}$. \\
        \vspace{-0.4cm}
        \begin{algorithmic}[1]
            \STATE Init mini-batch $B = \O $.
            \FOR{$i=1, \dots, \frac{N}{2} $}
                \STATE Sample random image $x$.
                \STATE Generate positive pair $[x, x']$ of $x$.
                \STATE Randomly select global hard negative $x_{\text{hn}}$ of $x$ from \texttt{KNN}.
                \STATE Generate positive pair $[x_\text{hn}, x_\text{hn}']$ of $x_\text{hn}$.
                \STATE Update mini-batch $B = B \cup \{ [x, x'], [x_\text{hn}, x_\text{hn}'] \} $
            \ENDFOR
        \end{algorithmic}
        \end{algorithm}

    \subsection{Local Crops Ensembling}
        \label{sec:lce}
        In copy detection, there are numerous edited copy instances of small regions, which present significant challenges to algorithms. In response to this issue, a straightforward approach is local crops ensembling, abbreviated as LCE. The primary concept of LCE is to crop query $q$ and reference $r$ according to a fixed rule, followed by pairwise comparisons between the cropped queries and references. Ultimately, LCE takes the maximum score as the copy score of query $q$ and reference $r$. For matcher, LCE takes the highest copy probability, while for descriptor, it takes the maximum cosine similarity.
        
        As posted in Tab.~\ref{tab:comparison_with_sotas}, LCE significantly enhances matcher performance (2.9\%+ $\mu$AP / 4.0\%+ RP90). However, the downside of LCE is also evident: it requires substantial computational resources. In our implementation, we cropped 26 regions (25 local + 1 global) from query and 10 regions (9 local + 1 global) from reference, leading to a corresponding computation cost of 260x. These regions include rotation and different ratios of crops. Please refer to our code for more details of LCE. It is important to emphasize that we design this complexity mainly because some approaches in Tab.~\ref{tab:comparison_with_sotas} adopt sophisticated rules for the best performance and we follow them for fair comparison.

\section{More about Candidates List for Matcher}
    \label{sec:candidate}
    As illustrated in Tab.~\ref{tab:model_property}, matcher can only perform classification pair by pair. According to Tab.~\ref{tab:disc21_datasets}, \texttt{reference set} has 1M images and \texttt{dev set} or \texttt{test set} has over 25k images. If we force matcher to classify all possible query and reference pairs, it will incur exceptionally high computational cost. To solve this problem, descriptor is first utilized to recall as many edited copy cases as possible to generate a candidate list. Next, we apply matcher to classify all pairs within this candidate list. When recalling with descriptor, LCE mentioned in Sec.~\ref{sec:lce} is employed. 
    
    Specifically, for each crop of query, we identify $k$-NN in reference set and each query recalls 390 candidates. We ensemble the candidate lists from ``Baseline'' (ViT-B, $224 \times 224$) and ``CopyNCE'' (ViT-B, $224\times224$), recalling a total of 780 candidates. And then duplicated candidates for each query will be removed and the top 400 candidates form a multi-model fused candidate list. Finally, we use matcher ``CopyNCE'' (ViT-S, $224\times224$) in conjunction with LCE to select the top 10 highest-scoring candidates to generate the final candidate list. All matcher will subsequently utilize this candidate list for inference. Recall of different steps is listed in Tab.~\ref{tab:recall_of_dev_candidates}. Note that recall could be viewed as the upper bound of mAP and $\mu$AP, \textit{i.e.,} our reported $\mu$AP of matcher in Tab.~\ref{tab:comparison_with_sotas} can no longer be greater than 93.4\%. More details could be found in our code.

    \begin{table*}[h]
        \begin{center}
        \begin{tabular}{c|cccc|ccc}
        \hline
                
        \hline\hline
        \multirow{2}{*}{Step} & \multicolumn{4}{c|}{Model} & \multirow{2}{*}{Tricks} & \multirow{2}{*}{\# Candidates} & \multirow{2}{*}{Recall} \\

        & Type & Method & Backbone & Resolution &  &  &  \\
        \hline
        
        0 & \multicolumn{4}{c|}{-} & - & 1,000,000 & 100.0\% \\
        \hline
        
        \multirow{2}{*}{1} & \multirow{2}{*}{Descriptor} & Baseline & \multirow{2}{*}{ViT-B} & \multirow{2}{*}{$224 \times 224$} & \multirow{2}{*}{LCE} & \multirow{2}{*}{400} & \multirow{2}{*}{94.3\%} \\

        &  & CopyNCE &  &  &  &  &  \\
        \hline
        
        2 & Matcher & CopyNCE & ViT-S & $224 \times 224$ & LCE & 10 & 93.4\% \\
                
        \hline\hline
        
        \hline
        \end{tabular}
        \end{center}
        \caption{\textbf{Recall of dev set part II in different candidates retrieval steps.} } 
        
        \label{tab:recall_of_dev_candidates}
    \end{table*}
    
\section{Solutions of DISC21 Phase 2}

    \begin{table*}[h]
        \begin{center}
        \begin{tabular}{c|ccccc|ccc}
        \hline
                
        \hline\hline
        \multirow{2}{*}{Step} & \multicolumn{5}{c|}{Model} & \multirow{2}{*}{Tricks} & \multirow{2}{*}{\# Candidates} & \multirow{2}{*}{Recall} \\

        & Type & Method & Pre-train & Backbone & Resolution &  &  &  \\
        \hline
        
        0 & \multicolumn{5}{c|}{-} & - & 1,000,000 & 100.0\% \\
        \hline
        
        \multirow{2}{*}{1} & \multirow{2}{*}{Descriptor} & \multirow{2}{*}{Finetune} & Baseline & \multirow{2}{*}{ViT-B} & \multirow{2}{*}{$224 \times 224$} & \multirow{2}{*}{LCE} & \multirow{2}{*}{400} & \multirow{2}{*}{91.3\%} \\

        &  &  & CopyNCE &  &  &  &  &  \\
        \hline
        
        2 & Matcher & Finetune & CopyNCE & ViT-S & $224 \times 224$ & LCE & 10 & 90.1\% \\
                
        \hline\hline
        
        \hline
        \end{tabular}
        \end{center}
        \caption{\textbf{Recall of test set in different candidates retrieval steps.} } 
        
        \label{tab:recall_of_test_candidates}
    \end{table*}

    DISC21 Phase 2 used \texttt{test set} to measure performance, which is more challenging compared to \texttt{dev set part II}. To achieve better results on \texttt{test set}, we follow the official rules and finetune both descriptor and matcher on \texttt{dev set part I}, with the tuning parameters detailed in Tab.~\ref{tab:all_finetune_detailed_settings}. It is important to note that, according to DISC21 rules, ``Descriptor Track'' only allows features with 256 dim. Therefore, we reduced the default linear head dimensions from 512 to 256.

    For ``Descriptor Track'', we finetune ``CopyNCE'' descriptor (ViT-B, $336\times336$) to obtain the final descriptor and perform inference at $336\times336$. The result presented in Tab.~\ref{tab:leaderboard_of_phase_2} is achieved with feature stretching (as described in Sec.~\ref{sec:feature_stretching}). To obtain the candidate list for matcher, we repeat the process outlined in Sec.~\ref{sec:candidate} with the distinction that descriptors used for retrieval are finetuned on \texttt{dev set part II}. The recall after different steps are listed in Tab.~\ref{tab:recall_of_test_candidates}. Finally, we utilized finetuned matcher (ViT-S, $336\times336$) to achieve the result shown in Tab.~\ref{tab:leaderboard_of_phase_2}.

\section{More about Reverse Operation of Table}
    \begin{figure}[ht]
        \begin{center}
            \includegraphics[width=1.0 \linewidth]{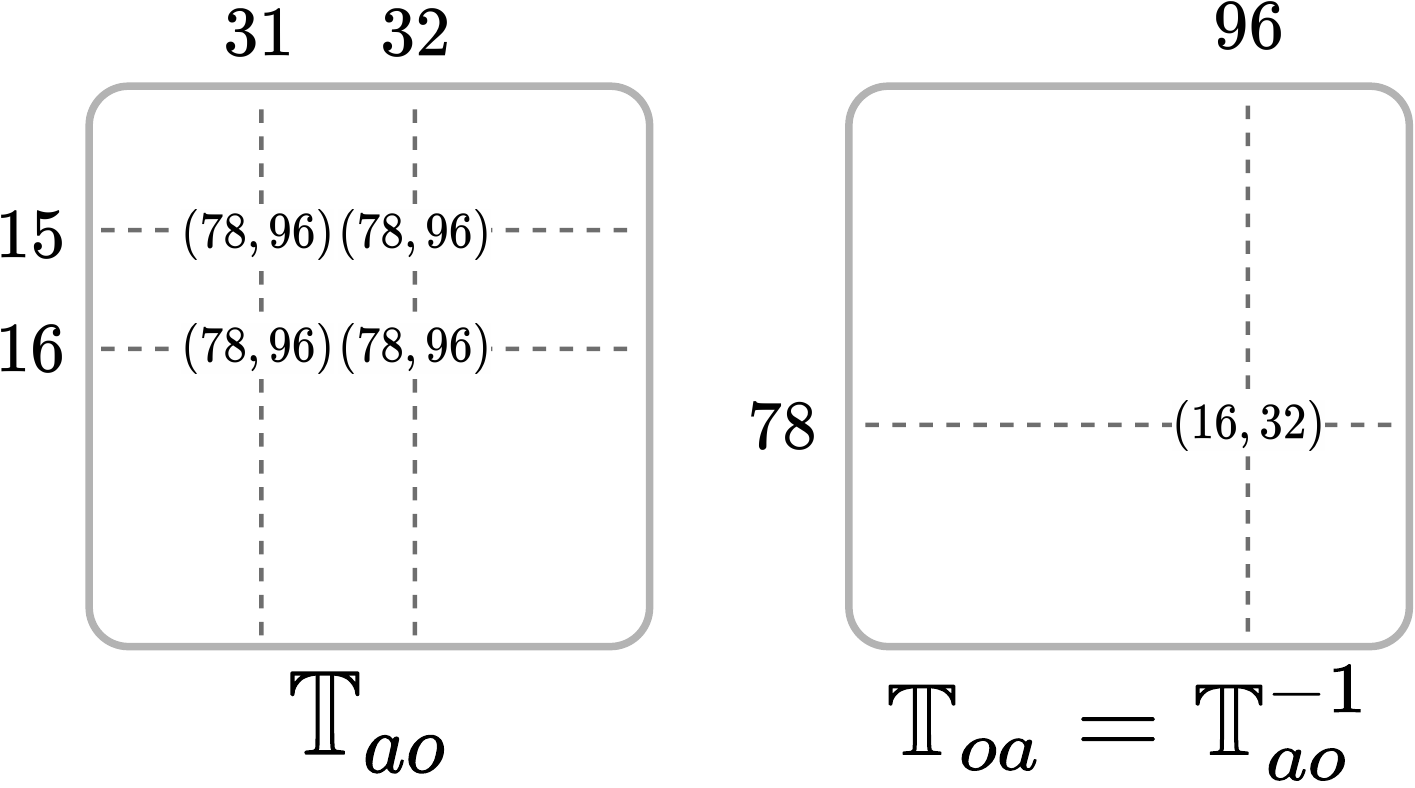}
        \end{center}
        \caption{\textbf{Reverse issue of table $\mathbb{T}$.}}
        \label{fig:reverse_issue}
    \end{figure}
    When performing the reverse operation on the coordinate table $\mathbb{T}$, the keys and values are often not uniquely corresponding, as multiple keys may correspond to the same value. If table $\mathbb{T}$ is reversed, a single key may correspond to multiple values. In such cases, we apply a row-by-row reversal rule, where the subsequently reversed value overrides the previous one. For instance, in the case shown in the diagram, if image $\text{I}_o$ is magnified to twice its size and overlaid on image $\text{I}_a$, the coordinates (15, 31), (15, 32), (16, 31), and (16, 32) in image $\text{I}_a$ are tracked to the pixel at coordinate (78, 96) in image $\text{I}_o$. Therefore, if table $\mathbb{T}_{ao}$ is reversed, key coordinate (78, 96) will correspond to multiple value coordinates. Following the row-by-row reversal process, the later value will override the earlier one, making (78, 96) correspond to (16, 32).

\section{More about ResNet experiments}
    \label{sec:resnet_discussion}
    For a $224 \times 224$ input image, ResNet-50 yields $7 \times 7 = 49$ regional features, whereas ViT with the default patch size of $16 \times 16$ produces $14 \times 14 = 196$ patch tokens. To align the spatial granularity of ResNet-50 with that of ViT, we upsample the input to $448 \times 448$ in our ResNet-50 experiments. Without this alignment, CopyNCE fails to confer performance gain. We attribute this phenomenon to the fact that, for descriptor, the fraction of image area occupied by each regional feature critically modulates the efficacy of CopyNCE, and $14 \times 14$ appears to constitute an optimal trade-off. Tab.~\ref{tab:scaling_of_descriptor} also corroborates this hypothesis that increasing the input resolution from $224 \times 224$ to $336 \times 336$ does not improve performance.

\section{More Visualization}
\begin{figure*}[h]
    \begin{center}
        \includegraphics[width=1.0 \linewidth]{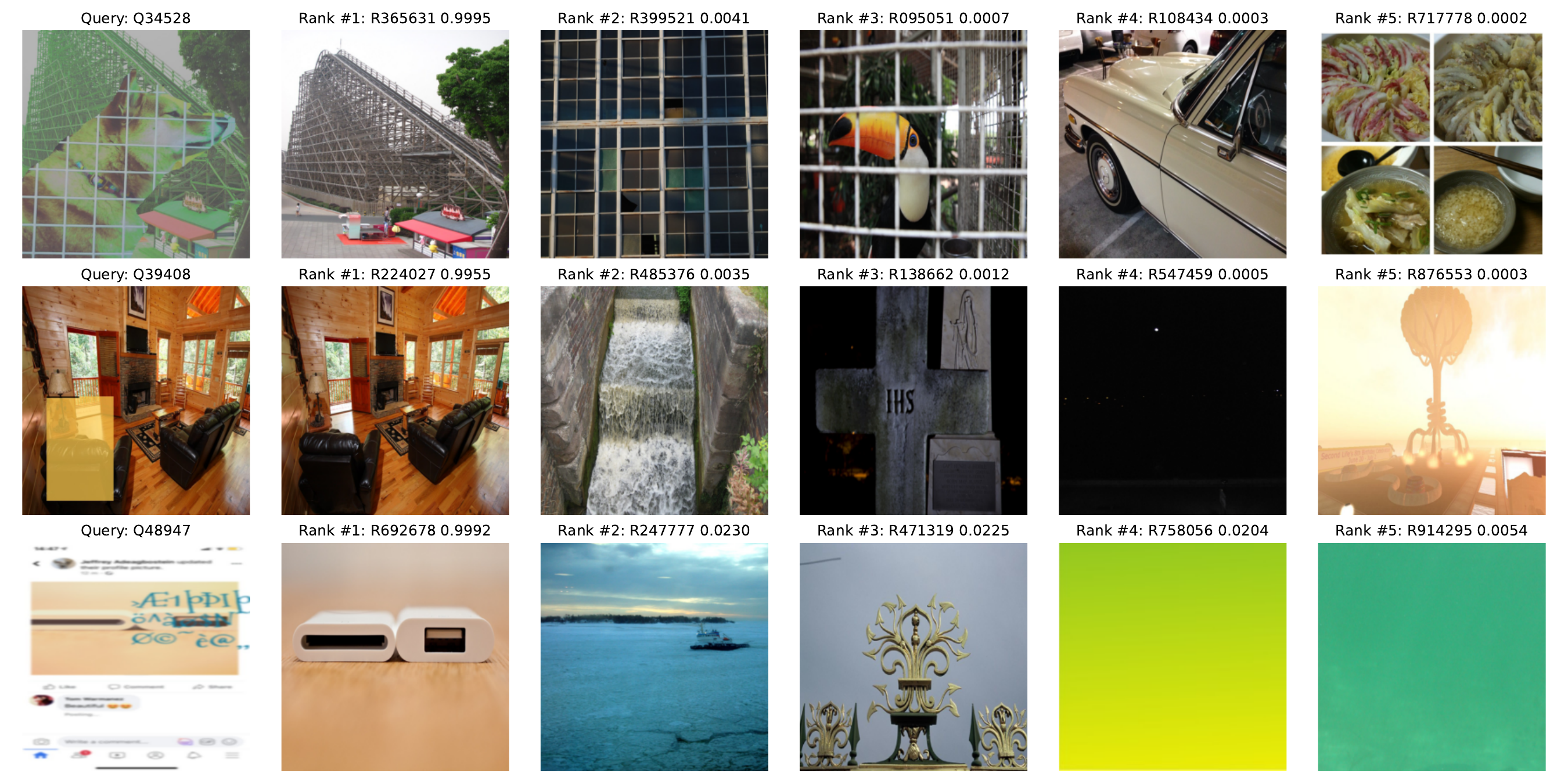}
    \end{center}
    \caption{\textbf{Success cases of matcher.} In each row, the query image is placed in the first column, followed by five recalled reference images sorted by score in descending order. Each image is titled with its rank, ID and copy score. Notably, the ground truth is always ranked first.}
    \label{fig:matcher_success}
\end{figure*}

\begin{figure*}[h]
    \begin{center}
        \includegraphics[width=1.0 \linewidth]{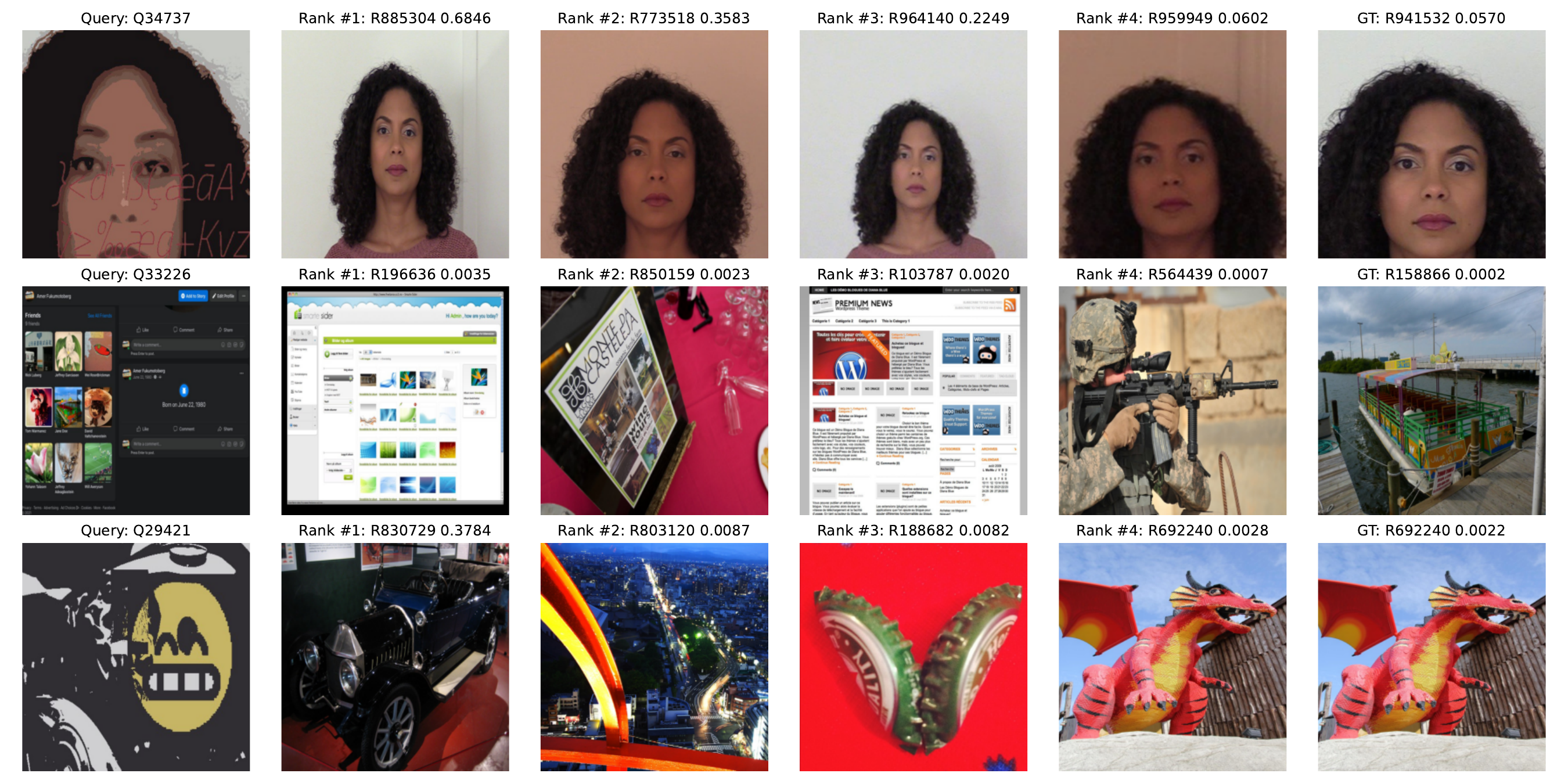}
    \end{center}
    \caption{\textbf{Failure cases of matcher.} In each row, the query image is placed in the first column, followed by four recalled reference images sorted by score in descending order. Each image is labeled with its rank, ID and copy score. In these cases, the ground truth copied image is placed last.}
    \label{fig:matcher_failure}
\end{figure*}

\begin{figure*}[h]
    \begin{center}
        \includegraphics[width=1.0 \linewidth]{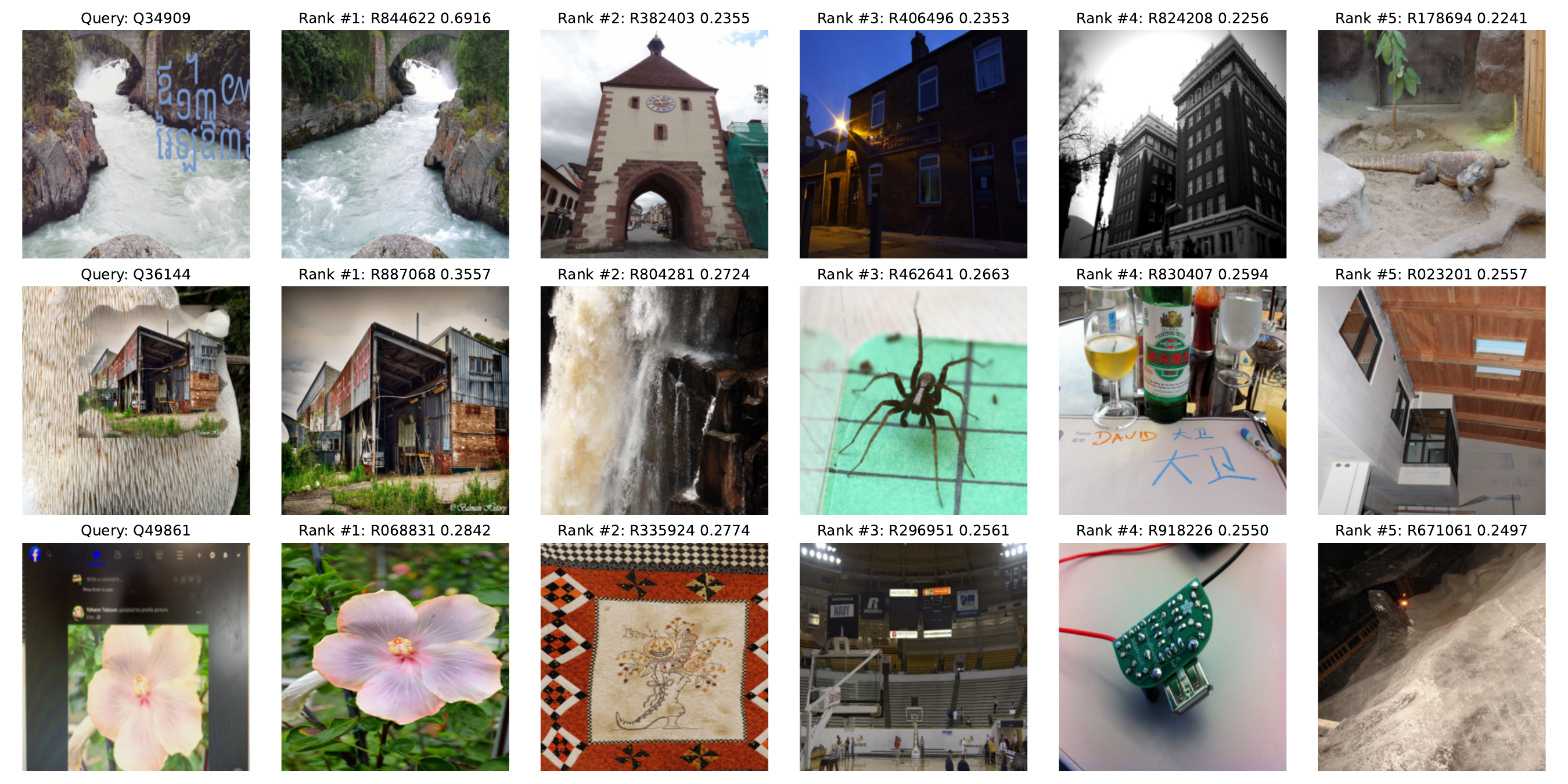}
    \end{center}
    \caption{\textbf{Success cases of descriptor.} In each row, the query image is placed in the first column, followed by five recalled reference images sorted by score in descending order. Each image is titled with its rank, ID and copy score. Notably, the ground truth is always ranked first.}
    \label{fig:descriptor_success}
\end{figure*}

\begin{figure*}[h]
    \begin{center}
        \includegraphics[width=1.0 \linewidth]{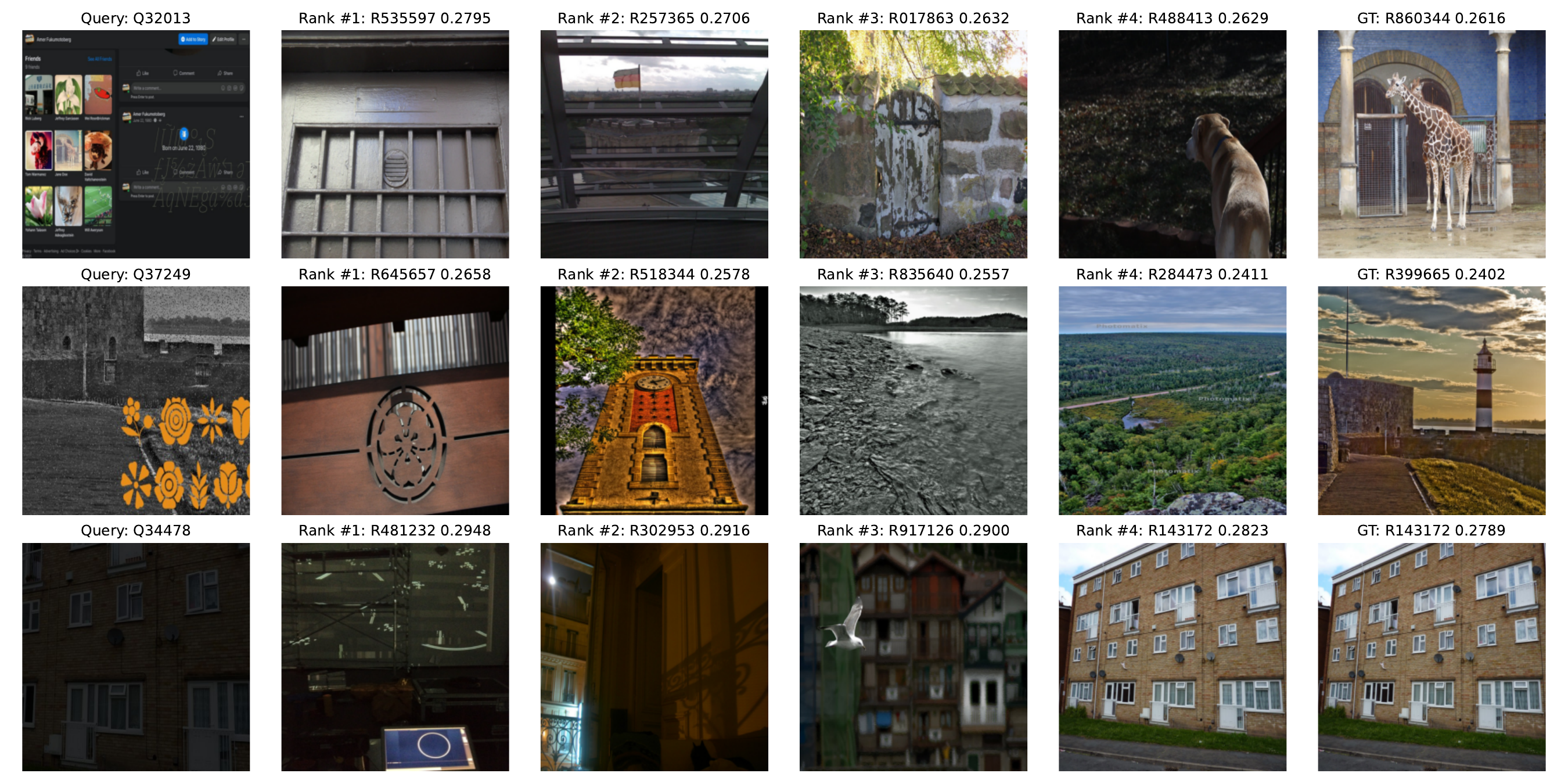}
    \end{center}
    \caption{\textbf{Failure cases of descriptor.} In each row, the query image is placed in the first column, followed by four recalled reference images sorted by score in descending order. Each image is labeled with its rank, ID and copy score. In these cases, the ground truth copied image is placed last.}
    \label{fig:descriptor_failure}
\end{figure*}

    To better visualize the effectiveness of CopyNCE series, we provide additional cases that include both success (Fig.~\ref{fig:matcher_success} for matcher and Fig.~\ref{fig:descriptor_success} for descriptor) and failure (Fig.~\ref{fig:matcher_failure} for matcher and Fig.~\ref{fig:descriptor_failure} for descriptor) examples of matcher and descriptor. In success cases, the copied images are correctly identified with the highest scores. Some of these cases even applied a large number of image transformations. This indicates that CopyNCE series is capable of handling most common edited copy cases. However, it remains evident that when copy area is relatively small, the descriptor is less effective compared to the matcher, as scores of descriptor decline significantly when the proportion of copy regions decreases. In failure cases, edited images generally employed highly exaggerated transformations that fused the model, which still remained a challenge to our models.

\section{Potential Limitations}
    Unlike image-level supervision, CopyNCE requires a coordinate table as their supervisory signal. For an edited copy pair under the default settings, a table containing $224 \times 224$ key-value pairs needs to be generated. Therefore, training process may encounter CPU bottlenecks, resulting in longer training time for CopyNCE compared to that for image-level supervised pipelines.

    Besides, to ensure a fair comparison with other SOTA methods, CopyNCE utilizes the LCE trick when obtaining matcher results, which significantly enhances performance. However, in practical applications, the high computational cost of LCE makes it nearly impossible to be leveraged in real-life applications. Regarding this limitation, we argue that even without LCE, CopyNCE still is still capable to yield competitive results, as shown in some cases in Tab.~\ref{tab:comparison_with_sotas}.
    
    Finally, since copy detection emphasizes features that contain rich texture information, such features often lack semantic understanding compared to models of SSL or other image retrieval tasks. Consequently, they tend to focus on detailed texture information between two images while overlooking semantic foregrounds.


\end{document}